\documentclass{article} 
\usepackage{iclr2022_conference,times}

\usepackage{amsmath,amsfonts,bm}

\def\eqref#1{equation~\ref{#1}}

\def\1{\bm{1}}

\DeclareMathAlphabet{\mathsfit}{\encodingdefault}{\sfdefault}{m}{sl}
\SetMathAlphabet{\mathsfit}{bold}{\encodingdefault}{\sfdefault}{bx}{n}

\DeclareMathOperator*{\argmax}{arg\,max}
\DeclareMathOperator*{\argmin}{arg\,min}

\usepackage{hyperref}
\usepackage{url}

\usepackage{wrapfig}
\usepackage{microtype}
\usepackage{graphicx}
\usepackage{subfigure}
\usepackage{booktabs} 

\usepackage{hyperref}

\usepackage{amsmath}
\usepackage{amssymb}
\usepackage{mathtools}
\usepackage{amsthm}
\usepackage{amsmath}
\usepackage{amssymb}
\usepackage{multirow}
\usepackage{xcolor}
\newcommand{\ourAcronym}{MetaMD}
\newcommand{\ourModel}{Meta Mirror Descent}

\newcommand{\cut}[1]{}

\newcommand{\imgrad}{h}
\newcommand{\sloss}{\mathcal{L}_{tr}}
\newcommand{\keypoint}[1]{\noindent\textbf{#1}\quad}

\RequirePackage{algorithm}
\RequirePackage{algorithmic}
\usepackage[capitalize,noabbrev]{cleveref}

\theoremstyle{plain}
\newtheorem{theorem}{Theorem}[section]

\theoremstyle{definition}

\theoremstyle{remark}

\usepackage[textsize=tiny]{todonotes}

\title{Meta Mirror Descent: Optimiser Learning for Fast Convergence}

\iclrfinalcopy
\author{Boyan Gao, Henry Gouk \\
University of Edinburgh\\
\texttt{\{boyan.gao, henry.gouk\}@ed.ac.uk} \\
\And
Hae Beom Lee  \\
KAIST, Suth Korea\\
\texttt{\{haebeom.lee\}@kaist.ac.kr} \\
\And
Timothy M. Hospedales  \\
University of Edinburgh\\
Samsung AI Research, Cambridge\\
\texttt{\{t.hospedales\}@ed.ac.uk} \\
}

\begin{document}

\maketitle

\begin{abstract}
Optimisers are an essential component for training machine learning models, and their design influences learning speed and generalisation. Several studies have attempted to learn more effective gradient-descent optimisers via solving a bi-level optimisation problem where generalisation error is minimised with respect to optimiser parameters. However, most existing optimiser learning methods are intuitively motivated, without clear theoretical support. We take a different perspective starting from mirror descent rather than gradient descent, and meta-learning the corresponding Bregman divergence. Within this paradigm, we formalise a novel meta-learning objective of minimising the regret bound of learning. The resulting framework, termed \ourModel{} (\ourAcronym{}), learns to accelerate optimisation speed. Unlike many meta-learned optimisers, it also supports convergence and generalisation guarantees and uniquely does so without requiring validation data. We evaluate our framework on a variety of tasks and architectures in terms of convergence rate and generalisation error and demonstrate strong performance.
\end{abstract}
\section{Introduction}

Gradient-based optimization algorithms, such as stochastic gradient descent (SGD), are fundamental building blocks of many machine learning algorithms -- notably those focused on training linear models and deep neural networks.
These methods are typically developed to solve a broad class of problems, and therefore the method developers make as few assumptions about the target problem as possible. This leads to a variety of general purpose techniques for optimization, but such generality often comes with slower convergence. By taking advantage of more information about the target problem, one is typically able to design more efficient---but less general---optimization algorithms. For example, by taking advantage of second order information, Newton's method is able to converge to optima in many fewer iterations than gradient descent, which uses only first order information. However, the application of Newton's method is limited compared to gradient descent, because it can only be used to solve problems where the second order information exists and can be computed efficiently. Another challenge in a non-convex deep learning context, is that that many of the empirically fastest optimizers such as Adam \citep{kingma2014adam} lack convergence guarantees. 

%
While one line of research hand-designs optimisers to exploit known properties of a particular problems, a complementary line of research focuses on situations where optimisation problems come in families. This allows using meta-learning techniques to fit an optimiser to the given problem family with the goal of maximising convergence speed or generalisation performance. For example, in the many-shot regime,  \citet{andrychowicz2016learning} and \citet{wichrowska2017learned} learn black-box neural optimisers to accelerate training of neural networks, while \citet{bello2017neural} learn symbolic gradient-based optimisers to improve generalisation. MAML \citep{finn2017model} and Meta-SGD~\citep{li2017meta} learned   initialisation and learning rate for SGD training of neural networks with good generalisation performance in the few-shot regime. Later generalisations focused on learning problem family-specific curvature information \citep{park2019meta,flennerhag2019meta}. Nevertheless, most existing learned optimisers such as \cite{andrychowicz2016learning,wichrowska2017learned,flennerhag2019meta,bello2017neural} can not provide convergence or generalisation guarantees. 

In this work, we revisit the optimizer learning problem from the perspective of \emph{mirror descent}. Mirror descent introduces a Bregman divergence that regularises the distance between current and next iterate, introducing a strongly convex sub-problem that can be optimised exactly. In mirror descent, the choice of Bregman divergence determines optimisation dynamics. In a meta-learning context, the Bregman divergence thus provides a novel representation of an optimisation strategy that can be fit to a given family of optimisation problems, leading to our learned optimiser termed \ourModel{} (\ourAcronym{}). Existing learned optimisers do not have a formal notion of convergence rate, and in practice typically optimise a meta-objective reflecting training or validation loss after a fixed number of iterations. In contrast, \ourAcronym{} is directly trained to optimise the convergence rate bound for mirror descent. Importantly, this means we can adapt theoretical guarantees from mirror descent to provide convergence guarantees for \ourAcronym{}, {an important property not provided by most learned optimsers}, and many hand-designed optimisers widely used in deep learning. 

An important issue in meta-learning a mirror descent algorithm is specifying the family of Bregman divergences to learn. Meta-learning with general Bregman divergences leads to an intractable tri-level optimisation problem. Thus, we seek a family of divergences for which the innermost optimisation has a closed form solution. The chosen paramaterisation should be complex enough to exhibit interesting optimisation dynamics, simple enough to provide a closed form solution, while always providing a valid Bregman divergence. We provide an example parameterisation that meets all these desideratum in the form of a mixture of diagonal matrices. In contrast to methods such as Meta-SGD, Meta-Curvature, and WarpGrad, this means that the learned optimisation strategy is more expressive insofar as being state-dependent: It can change in different parts of the parameter-space.

Empirically we demonstrate that we can train \ourAcronym{} for fast convergence given a model architecture and a suite of training tasks. We then deploy it to novel testing tasks. On novel problems, \ourAcronym{} provides fast convergence compared to many existing hand-designed optimisers. 
\section{Related work} 
Meta-learning aims to extract some notion of `\emph{how to learn}' given a task or distribution of tasks \citep{hospedales2020meta}, such that new learning trials are better or faster. These two stages are often called meta-training, and meta-testing respectively. Key dichotomies include: meta-learning from a single task vs a task distribution; the type of meta-knowledge to be extracted; and long- vs short-horizon meta-learning. For few-shot problems with short optimization horizons, the seminal model-agnostic meta-learning (MAML) \cite{finn2017model} learns an initial condition from which only a few optimisation steps are required solve a new task. Meta-SGD \cite{li2017meta} and Meta-Curvature \cite{park2019meta} extend MAML by learning a parameter-wise learning rate, and a preconditioning curvature matrix respectively. 
Another group of methods focus on larger scale problems in terms of dataset size and optimization horizon. For example, 
neural architecture search (NAS) \cite{real2019regularized,zoph2016neural}  discovers effective neural architectures. MetaReg~\cite{balaji2018metareg} meta-learns  regularization parameters to improve domain generalisation. ARL \citep{Gao_2021_ICCV} meta-learns a loss function to improve robustness of learning form noisy labels. 

Several studies focus specifically on optimiser meta-learning for many-shot  problems, which we address here. In this case, the extracted meta-knowledge spans learning rates for SGD \citep{micaelli2021gradient},  symbolic gradient-descent rules \citep{bello2017neural}  neural network gradient-descent rules \citep{andrychowicz2016learning,li2017learnToOptimize}, and gradient-free optimisers 
\citep{sandler2021meta,chen2017l2lgdgd}. Differently to the gradient-descent based methods, we start from the perspective of mirror descent, where  mirror descent's Bregman Divergence provides an target for meta-learning. This perspective has several benefits, notably the ability to derive a learned optimizer with convergence and generalisation guarantees. While our framework is general, our practical instantiation for efficient implementation uses a divergence defined by a mixture of diagonal Malanobis distances. This can be interpreted as a mixture of element-wise learning rates for SGD, related to \cite{micaelli2021gradient}. However, we provide convergence guarantees, do not rely on a validation set, and demonstrate cross-dataset generalisation theoretically and empirically, enabling us to amortize meta-learning cost. In contrast, \cite{micaelli2021gradient}'s single task meta-learner needs to repeat meta-learning on each specific dataset to optimize per-dataset validation performance.

\section{Mirror Descent}
We formalise the problem of learning an optimiser using the Mirror Descent (MD) framework, which can be thought of as a generalisation of gradient descent. MD optimisers produce a series of progressively better estimates for the optimal parameters of the objective function. This is accomplished by solving a convex optimisation problem at each step, $t$,
\begin{align}
\label{eq:mirror-descent}
\theta_{t+1} = \argmin_{\theta} \langle \nabla_{\theta} \mathcal{L}_{tr}(\theta_t), \theta \rangle + \frac{1}{2 \eta} B_{\phi}(\theta || \theta_t)
\end{align}
where $\mathcal{L}_{tr}$ represents the training loss function, $\eta$ is the step size and $B_{\phi}$ denotes a Bregman divergence. Bregman divergences can be thought of as a way of measuring distance in parameter space, and each choice of Bregman divergence leads to a different optimisation algorithm. One can define Bregman divergences as
\begin{equation}
  B_{\phi}(\theta || \theta^\prime) = \phi(\theta) - \phi(\theta^\prime) - \langle \nabla \phi(\theta^\prime), \theta-\theta^\prime \rangle,
\end{equation}

\cut{
where $\phi$ is a $\lambda$-strongly convex function. If one chooses $\phi$ to be $\frac{1}{2}\|\cdot\|_2^2$, then after some simplification we obtain
\begin{equation}
    B_{\frac{1}{2}\|\cdot\|_2^2}(\theta||\theta^\prime) = \frac{1}{2}\|\theta - \theta^\prime\|_2^2,
\end{equation}
and the optimisation problem from Equation \ref{eq:mirror-descent} has a closed form solution,
\begin{equation}
    \theta_{t+1} = \theta_t - \eta \nabla_{\theta}\mathcal{L}_{tr}(\theta_t),
\end{equation}
which is the usual gradient descent update rule. There are several other choices of $\phi$ that result in existing algorithms specialised for various types of optimisation problems in machine learning. For example, choosing $\phi$ to be the negative entropy results in the Kullback-Leibler divergence, leading to the exponentiated gradient algorithm \citep{kivinen1997exponentiated}.
}
where $\phi$ is a $\lambda$-strongly convex function. There are several choices of $\phi$ that result in existing algorithms specialised for various types of optimisation problems in machine learning. For example, if one chooses $\phi$ to be $\frac{1}{2}\|\cdot\|_2^2$, then mirror descent becomes gradient descent, while choosing $\phi$ to be the negative entropy results in the Kullback-Leibler divergence leads to the exponentiated gradient algorithm \citep{kivinen1997exponentiated}.
A significant benefit of deriving new algorithms that fit into the mirror descent framework is that one can obtain a bound on the rate of convergence towards a minima $\theta_*$ for any valid choice of $\phi$. This bound also applies to learned divergences $B_\phi$.

\begin{theorem}
\label{regret_bound}
Let $B_{\phi}$ the Bregman divergence w.r.t $\phi: X \rightarrow \mathbb{R}$ and assume $\phi$ to be $\lambda$-strongly convex with respect to $||\cdot||$ in $\Theta$. Let $\Theta \subset X$. Set $\theta_1, \theta_* \in \Theta$ such that $\phi$ is differentiable in $\theta_1$. Then the following holds
\begin{align*}
    \sum_{t = 1}^T (l(\theta_t) - l(\theta_*)) & \leq \frac{B_{\phi}(\theta_* || \theta_1)}{\eta} + \frac{\eta}{2\lambda}\sum_{t= 1}^{T}||g_t||_*^2,
\end{align*}
where $\|\cdot\|_\ast$ is the dual norm of $\|\cdot\|$. $g_t$ represents the $t$ step gradient of the objective function $\ell$ whose minimiser is denoted as $\theta_*$.
\end{theorem}

\section{Meta-Learning a Mirror-Descent Bregman Divergence}
\subsection{Optimiser learning framework}
We propose a meta-learning algorithm to learn mirror descent optimisers. We consider the multi-task meta-learning setting \citep{hospedales2020meta,finn2017model}, assuming that a task distribution $p(\mathcal{T})$ available from from which we can draw tasks for meta-training, and that we will evaluate the learned optimizer by meta-testing on novel tasks from the same distribution. For gradient-based meta-learning, the meta-training procedure is conventionally framed as a bilevel optimization problem where the inner problem solves learning tasks given the optimiser, and the outer problem updates the optimiser \citep{hospedales2020meta}. The outer problem is to minimise some meta objective denoted $\mathcal{E}(\phi)$ with respect to the optimiser parameters. Since we are learning a mirror descent optimizer defined by a Bregman divergence $B_\phi$, this leads to a tri-level optimisation problem with a new layer corresponding to the problem given in Eq.~\ref{eq:mirror-descent} required to complete a single mirror descent step, 
\begin{align}
  \min_{\phi}&\, \mathcal{E}(\theta^\ast(\phi)) \label{eq:outer_loop} \\
  \text{s.t.}&\, \theta^\ast(\phi) = \argmin_{\theta} \mathcal{L}_{tr}(\theta) = (\pi_{\phi} \circ \pi_{\phi} ...  \circ \pi_{\phi})(\theta_1) \label{eq:inner_loop}\\
  \text{s.t.}& \, \pi_{\phi}(\theta_{t}) = \argmin_{\theta} \langle \nabla_{\theta} \mathcal{L}_{tr}(\theta_t), \theta \rangle + \frac{1}{2 \eta} B_{\phi}(\theta || \theta_t) \label{eq:mirror_loop}.
\end{align}
Solving the optimisation problem at each layer relies on the solution in other layers. In the outer loop (Eq.~\ref{eq:outer_loop}), the algorithm aims to learn a divergence by optimising the meta-objective $\mathcal{E}$, which evaluates the optimiser performance. To achieve this requires getting the best response from the mid-level problem (Eq.~\ref{eq:inner_loop}) where the base model is trained from the initialisation $\theta_0$ to $\theta_*$ by a sequence of $\pi_{\phi}$ which is the innermost problem in (Eq.~\ref{eq:mirror_loop})---which we denote the 
\emph{mirror loop}, due to the convention of using iterative solvers for such multi-level optimization problems. The mirror loop performs mirror descent updates using a Bregman divergence based on $\phi$. Compared with the standard bilevel problems in meta-learning, introducing this third layer adds significant cost to both meta-train and meta-test stages. However, with a suitable choice of divergence, we can obtain a closed-form solution for the mirror loop, which thus incurs similar cost to a standard bilevel optimisation problem. 
Our meta-learning framework for mirror descent optimisation is summarised in Alg~\ref{general_algorithm}.

\begin{algorithm}[t]
\caption{\ourModel{} learning algorithm.}
\label{general_algorithm}
\begin{algorithmic}[1]
\STATE {\bfseries Input: } $p(\mathcal{T}$), $\phi_M$ \hfill\COMMENT{Task distribution, and initial divergence paramaterised by $M$}
\STATE {\bfseries Output: } $\phi^*_M$
\WHILE{not converged or reached max steps}
    \STATE sample $T_1, ..., T_n$ from $p(\mathcal{T})$ \hfill\COMMENT{Get meta-train tasks}
    \FORALL{$T_i$}
      \STATE Init $\theta_i$ \hfill\COMMENT{Set random weights for base model}
        \STATE $\theta^{*}_{i} = \argmin_\theta \sloss (\theta_{i}, T_{i})$ \hfill\COMMENT{Train the base model}
        \STATE $\imgrad = \imgrad + \text{Hypergradient}(\sloss,\mathcal{E}, (\phi_M, \theta^*_i))$ \hfill\COMMENT{Obtain $d\mathcal{E}/dM$}
    \ENDFOR
    \STATE $M = M - \frac{\rho}{n} \imgrad  $ \hfill\COMMENT{Update the $\phi_M$ function}
\ENDWHILE
\end{algorithmic}
\end{algorithm}

\subsection{Divergence Parameterisation}
The paramaterisation of the divergence is important for a practical instantiation of our MetaMD framework. Ideally it should be expressive enough to represent interesting optimisation dynamics, while being simple enough to provide an efficient or closed form solution to the innermost convex mirror descent optimisation. We describe a reasonable compromise in the as follows.


Defining the parameterisation function of Bregman Divergence via the squared norm, $\phi(\theta) = \frac{1}{2}\theta^T M^2 \theta$, is a natural way to introduce a set of learnable parameters, $M$. We restrict $M$ to be a diagonal matrix, where we square the parameters to ensure positivity. One can also interpret this to be a parameter-wise learning rate. In this case, the mirror descent loop has a closed form solution,
\begin{align}
\theta_{t+1} = \theta_t - \eta M^{-2} \nabla_{\theta} \mathcal{L}_{tr}(\theta_{t}). \nonumber
\end{align}
The derivation of this closed-form mapping is given in Appendix~\ref{closed_inverse_mapping}. This paramaterisation is efficient, but provides limited capacity. To provide a better trade-off between capacity and efficiency, we can increase the capacity of $\phi$ by using $N$ diagonal matrices, ($M_1, ..., M_N$), and use a max operation to non-linearly aggregate the $N$ norms while  preserving convexity. The final form of $\phi$ is
\begin{align}
\phi(\theta)&= \max_{j \in \mathbb{N}_N} \, \theta^T M^2_j\theta,
\end{align}
which leads to a simple and efficient closed form solution,
\begin{align}
\theta_{t+1} &= \theta_t - \eta M^{-2}_{j^\ast} \nabla_{\theta} \mathcal{L}_{tr}(\theta_{t}) \\
j^\ast &= \argmax_{j \in \mathbb{N}_N} \, \theta_t^T M_j^2 \theta_t.
\end{align}

This setup provides increased expressivitiy through a mixture of learning rates, while also retaining a closed form solution for efficient mirror descent updates. A wide variety of better design choices are possible for $\phi$, but we focus on this simple mixture of diagonal matrices for the rest of this work.

\subsection{Meta-Objective} The next step is to define the meta-objective to optimise with respect to Bregman divergence $B_\phi$. An advantage of the mirror-descent framework is that we have a formal notion of convergence rate from an initialization $\theta_1$ to solution $\theta_*$ (from Theorem~\ref{regret_bound}). By defining the meta objective $\mathcal{E}(\phi)$ as a bound on the convergence rate, meta-learning $\phi$ leads to a faster optimiser. 

\cut{As mirror descent guarantees that it reaches the closest global minima from the initialisation in the distance measured by Bregman Divergence \citep{9488312}.}

In particular, we further bound Theorem~\ref{regret_bound}, by assuming that loss function is $L$-Lipschitz continuous, which bounds the gradient, leading to
\begin{align}
\sum_{t = 1}^T (l(\theta_t) - l(\theta_*)) \leq 
\frac{B_{\phi}(\theta_* || \theta_1)}{\eta} + \frac{\eta}{2\lambda}\sum_{t= 1}^{T}||g_t||_*^2 & \leq \frac{B_{\phi}(\theta_* || \theta_1)}{\eta} + \frac{\eta}{2\lambda}TL^2 \label{eq:regret_bb}.
\end{align}
Minimising the right hand side of Eq. \ref{eq:regret_bb} with respect to $\phi_M$ improves the convergence rate of the learned optimiser. As our goal is to optimise the expected speed of convergence on future tasks, we design a meta-objective that considers the average convergence rate over $n$ different meta-train tasks. Thus, we define the meta-objective as
\begin{align}
\mathcal{E}(\phi)= \frac{1}{n} \sum_{i=1}^n B_{\phi}(\theta_{T}^{(i)} || \theta_1^{(i)}) + \frac{k}{\lambda}, \label{eq:outer_obj} 
\end{align}
where $\theta_1^{(i)}$ and $\theta_T^{(i)}$ are the initial and final weights for task $i$, and we leave $k$ as a hyperparameter that can be tuned heuristically or using the relationship in Eq. \ref{eq:regret_bb}. \cut{Note that, in contrast to many meta-learning setups, the outer loop objective in Eq. \ref{eq:outer_obj} does not depend on any validation data.} The strong convexity parameter is given by $\lambda = \min_{i,j} M_{j,i,i}$. 

We solve the outer loop optimisation problem (Eq.~\ref{eq:outer_loop}) by gradient descent using $\partial\mathcal{E}(\phi_M)/\partial M$. As this gradient computation relies unrolling training trajectories in the inner loop where it is expensive to compute with standard reverse-mode differentiation. In this work we apply forward-mode  differentiation~\citep{franceschi2017forward} to solve this problem, as detailed in Appendix \ref{sec:hyper_gradient}.

\paragraph{Generalisation of the Learned Optimiser} Having meta-learned our MetaMD optimiser (Bregman divergence) on a set of training tasks, we can ask how well it is expected to perform on novel tasks? The training convergence rate, or outer objective value in 
Eq. \ref{eq:outer_obj} will be an optimistically biased estimate of the convergence rate one can expect on future tasks. Given the relatively simple family of Bregman divergences we employ, it is possible to construct a high-confidence bound on how biased this estimate will be, and therefore provide convergence guarantees that can be trusted:
\begin{theorem}
If we restrict $\|M_j\|_F \leq C$ and $\|\theta_1 - \theta_\ast\|_2 \leq r$, then the following holds with probability at least $1 - \delta$,
\begin{equation}
    \mathbb{E}\Bigg [ \sum_{t=1}^T (l(\theta_t) - l(\theta_\ast)) \Bigg ] \leq \frac{1}{n} \sum_{i=1}^n \frac{B_\phi(\theta_\ast^{(i)} || \theta_1^{(i)})}{\eta} + \frac{\eta TL^2}{2\lambda} + \frac{NC^2r^2}{2\sqrt{n}} + 3 \sqrt{\frac{Cr\text{ln}(2/\delta)}{8n}}.
\end{equation}\label{thm:generalisation}
\end{theorem}
This tells us that the expected convergence rate on novel tasks depends on the learning divergence on training tasks, plus complexity terms  such as the F-norm  of the meta-learned optimiser weights $M$. Note that restricting the diameter $r$ of the parameter space is usually required to obtain generalisation guarantees \citep{bartlett2017spectrally,long2020generalization, gouk2021distance}, so this is not an unusual or counterproductive requirement.

\section{Experiments}
We evaluate our learned optimiser and compare its performance on a variety of tasks against well-tuned standard baselines including SGD, SGD-M (SGD with momentum), Adam \citep{kingma2014adam} and RMSProp \citep{tieleman2012lecture}. We first explore synthetic tasks, followed by shallow neural networks on digit datasets, before finally evaluating training ResNet on CIFAR-10. 


\keypoint{Algorithm deployment pipeline:} For each set of experiments, we train \ourAcronym{} on a set of meta-train datasets, and evaluate it on a disjoint set of meta-test datasets. In the meta-test stage, models are trained by MetaMD (or competitors) using each dataset's standard training set, and evaluated on the corresponding test splits. We emphasize that for meta-testing, each optimiser consumes the same amount of data, and a comparable amount of compute per iteration. While MetaMD uses additional data and extra compute for the prior meta-training stage, this is a one-off that can be amortized across different meta-test problems of interest. We use $N=3$ diagonal matrices for MetaMD throughout the experiments.


\cut{
\begin{figure}[t]
\centering
\includegraphics[width=0.24\linewidth]{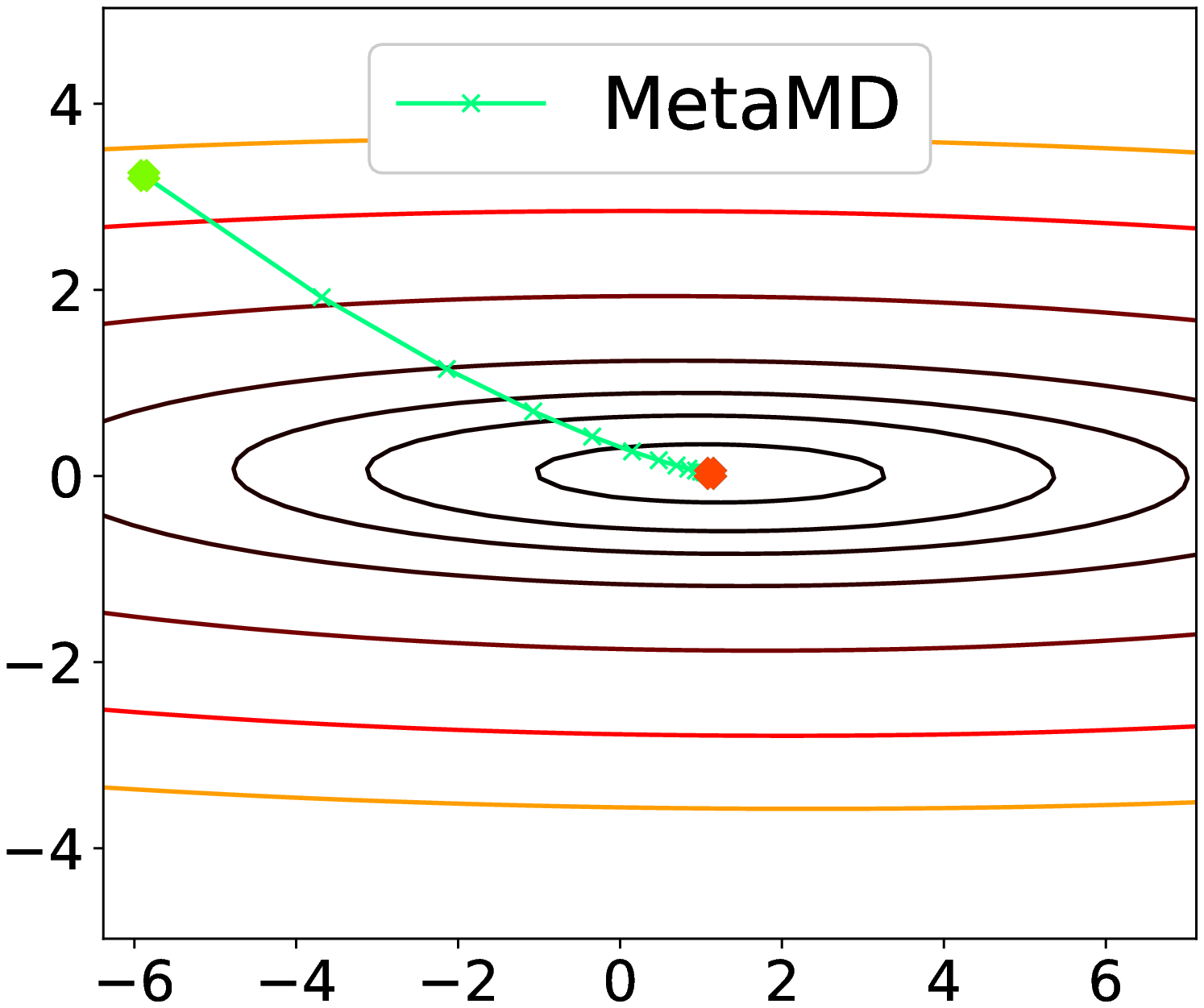}
\includegraphics[width=0.24\linewidth]{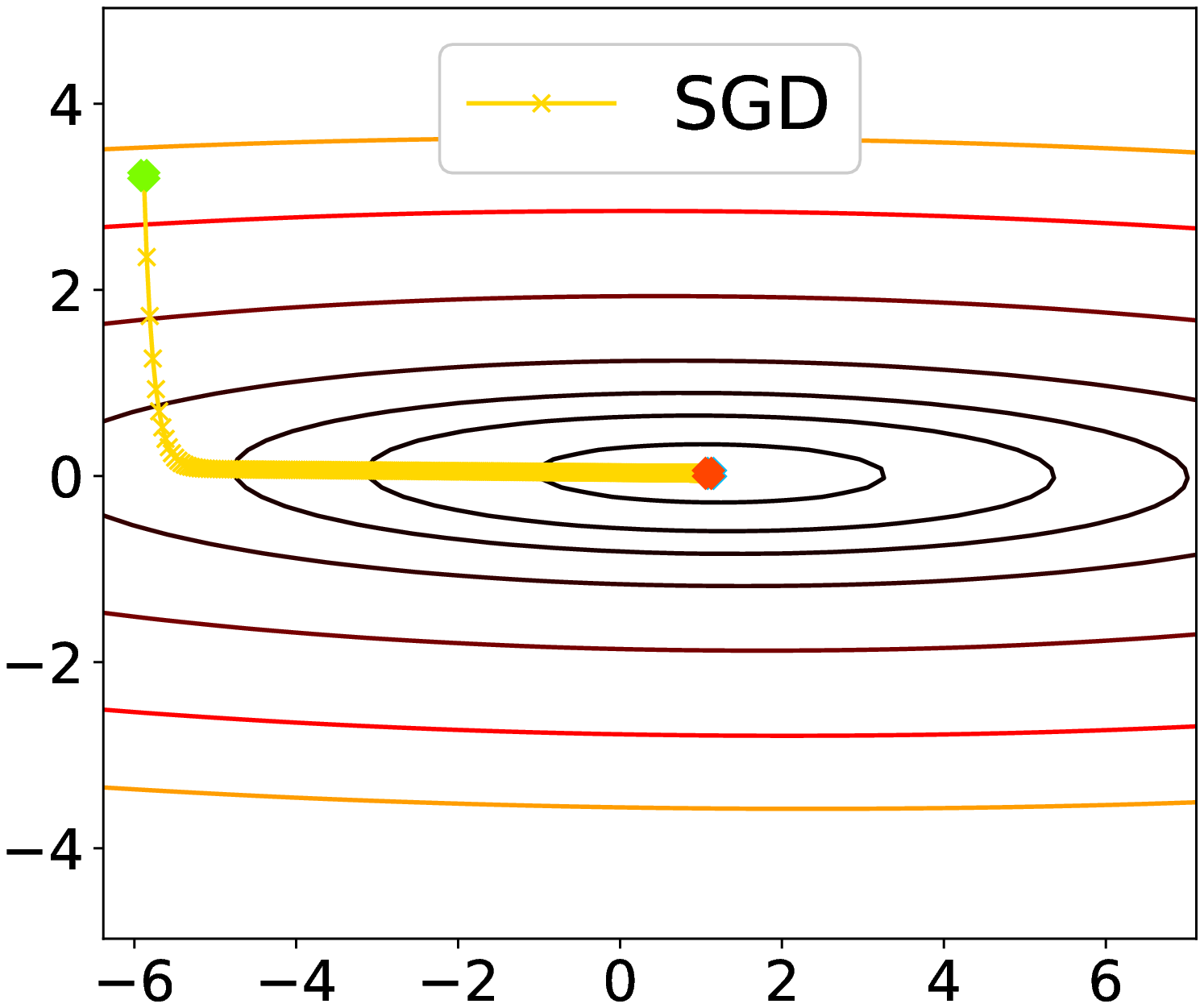}
\includegraphics[width=0.24\linewidth]{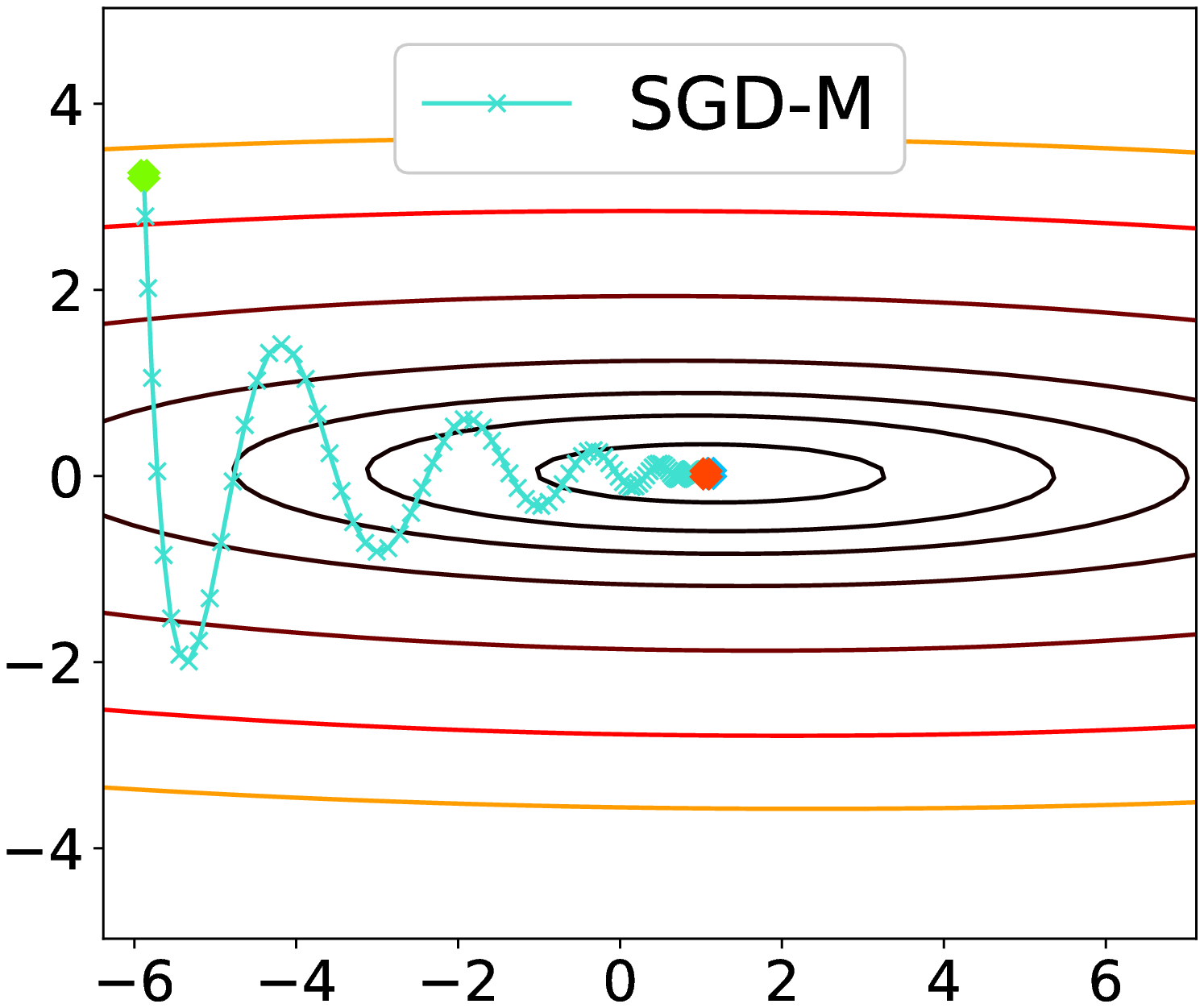}
\includegraphics[width=0.24\linewidth]{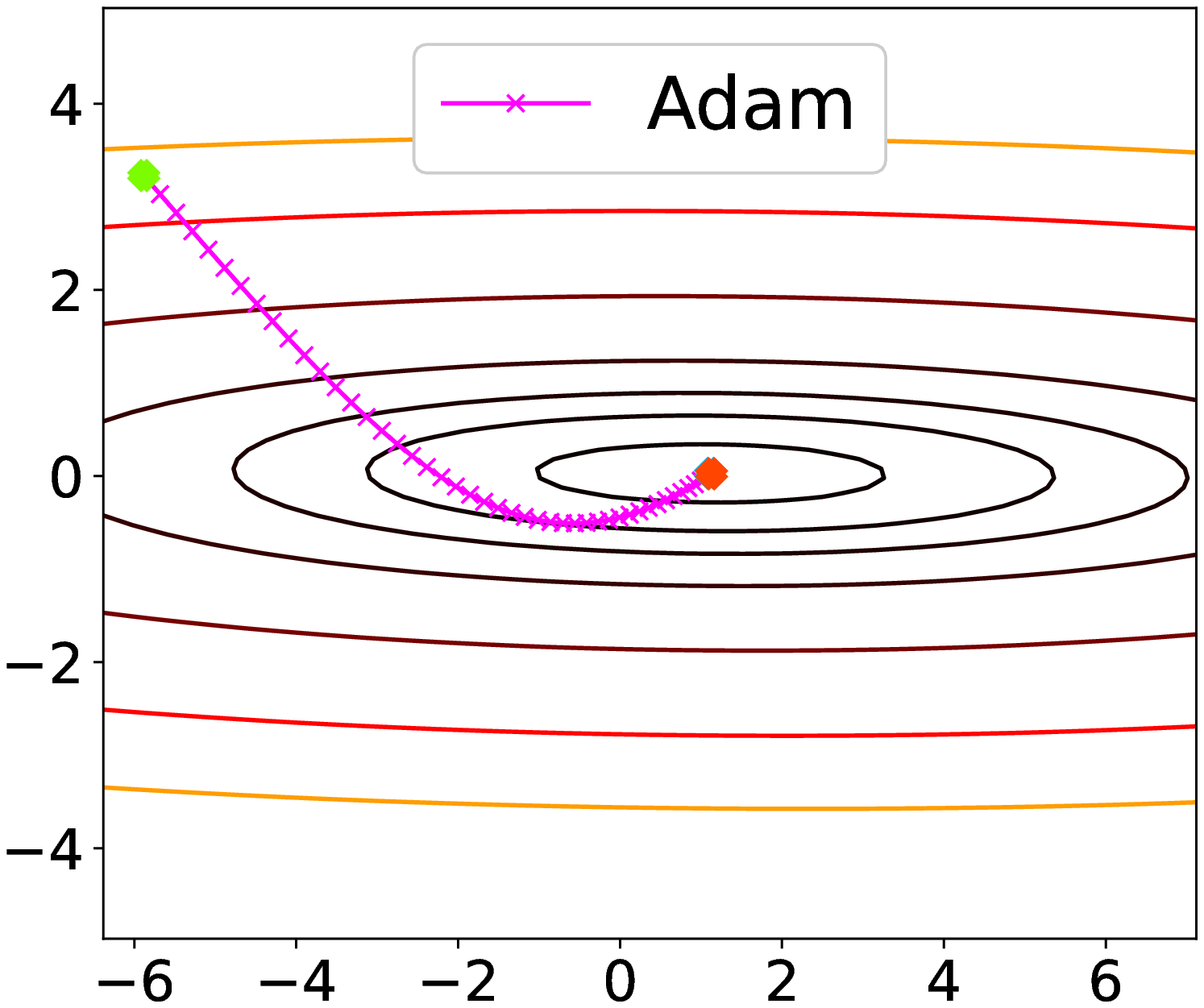}
\includegraphics[width=0.24\linewidth]{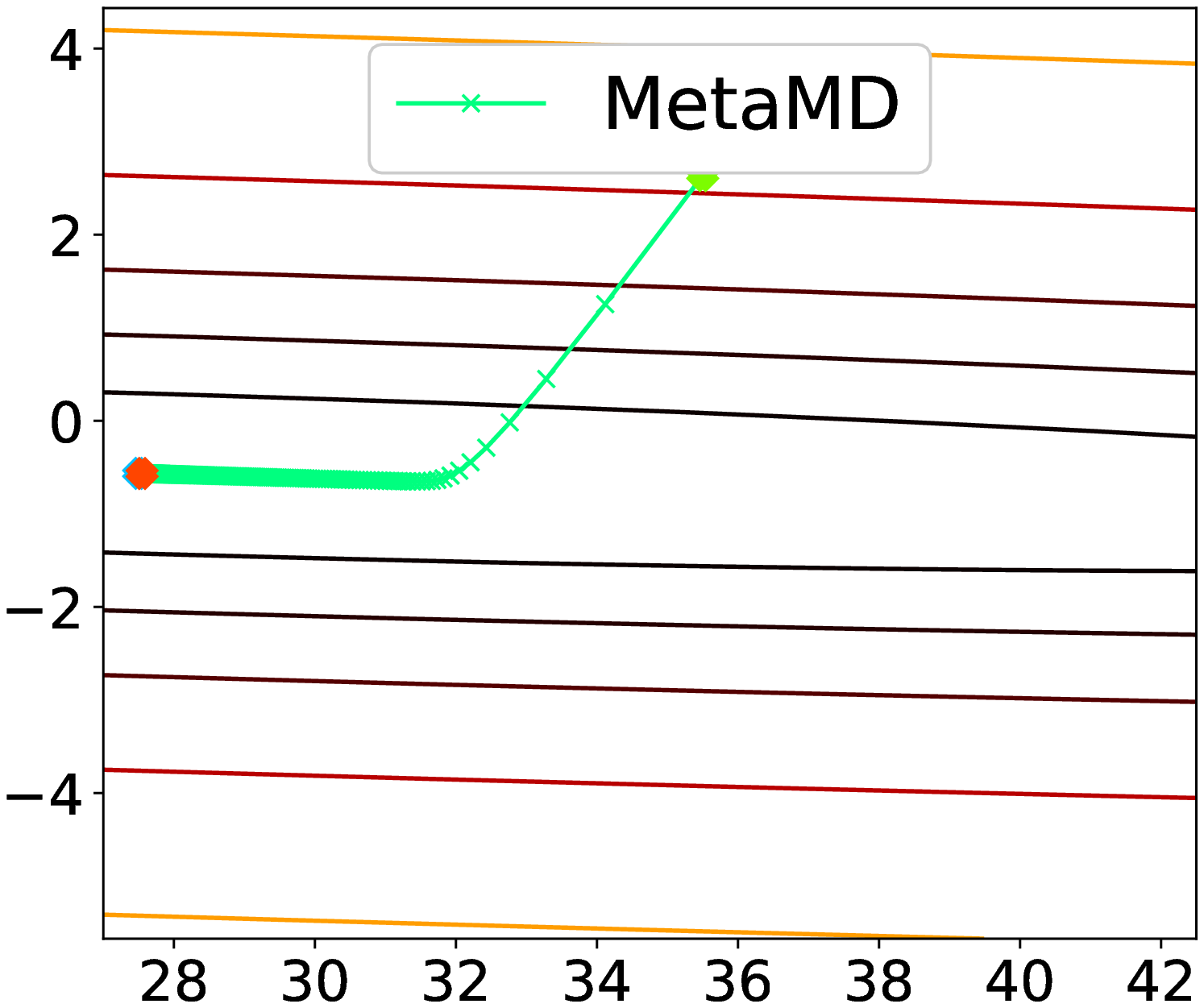}
\includegraphics[width=0.24\linewidth]{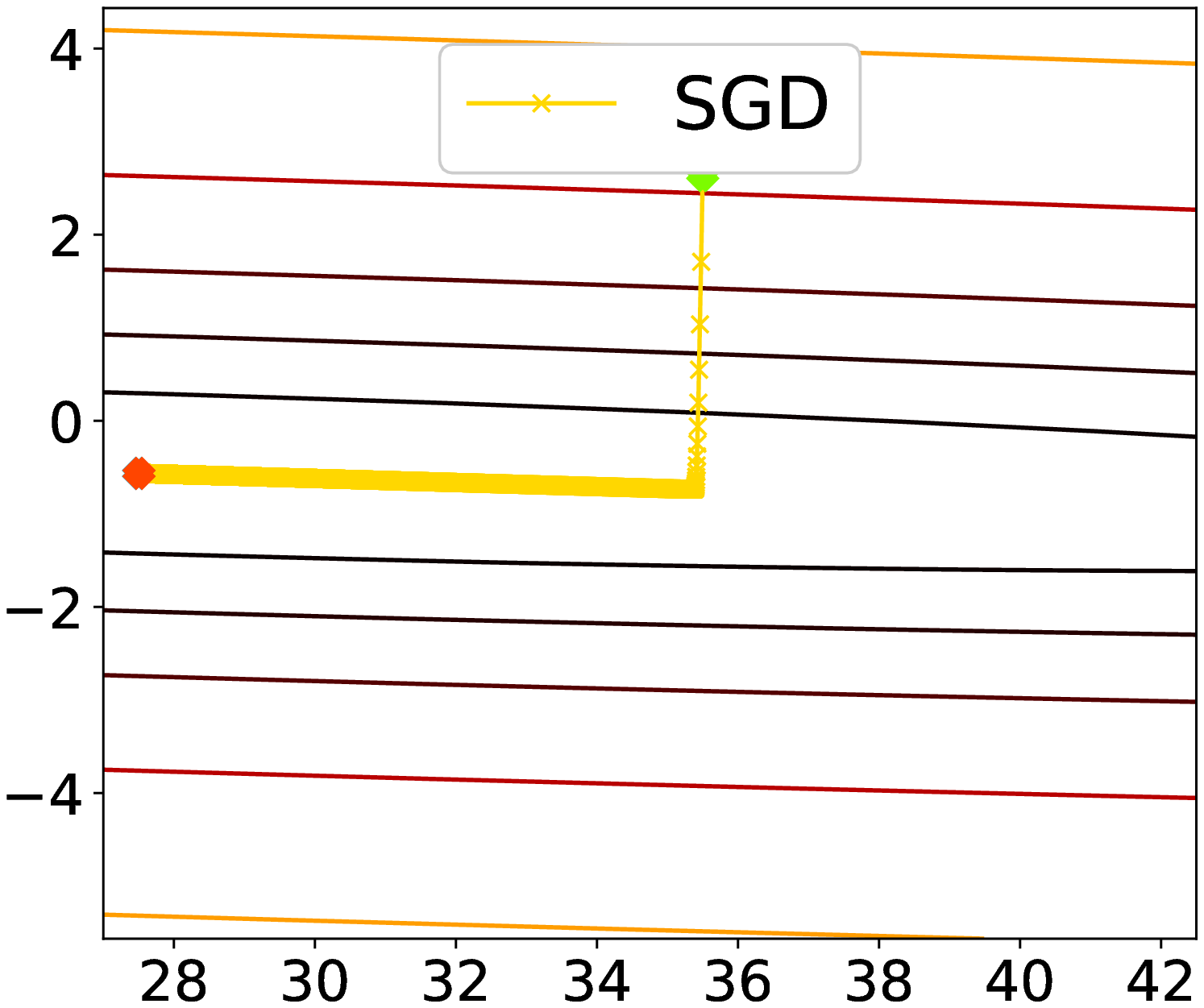}
\includegraphics[width=0.24\linewidth]{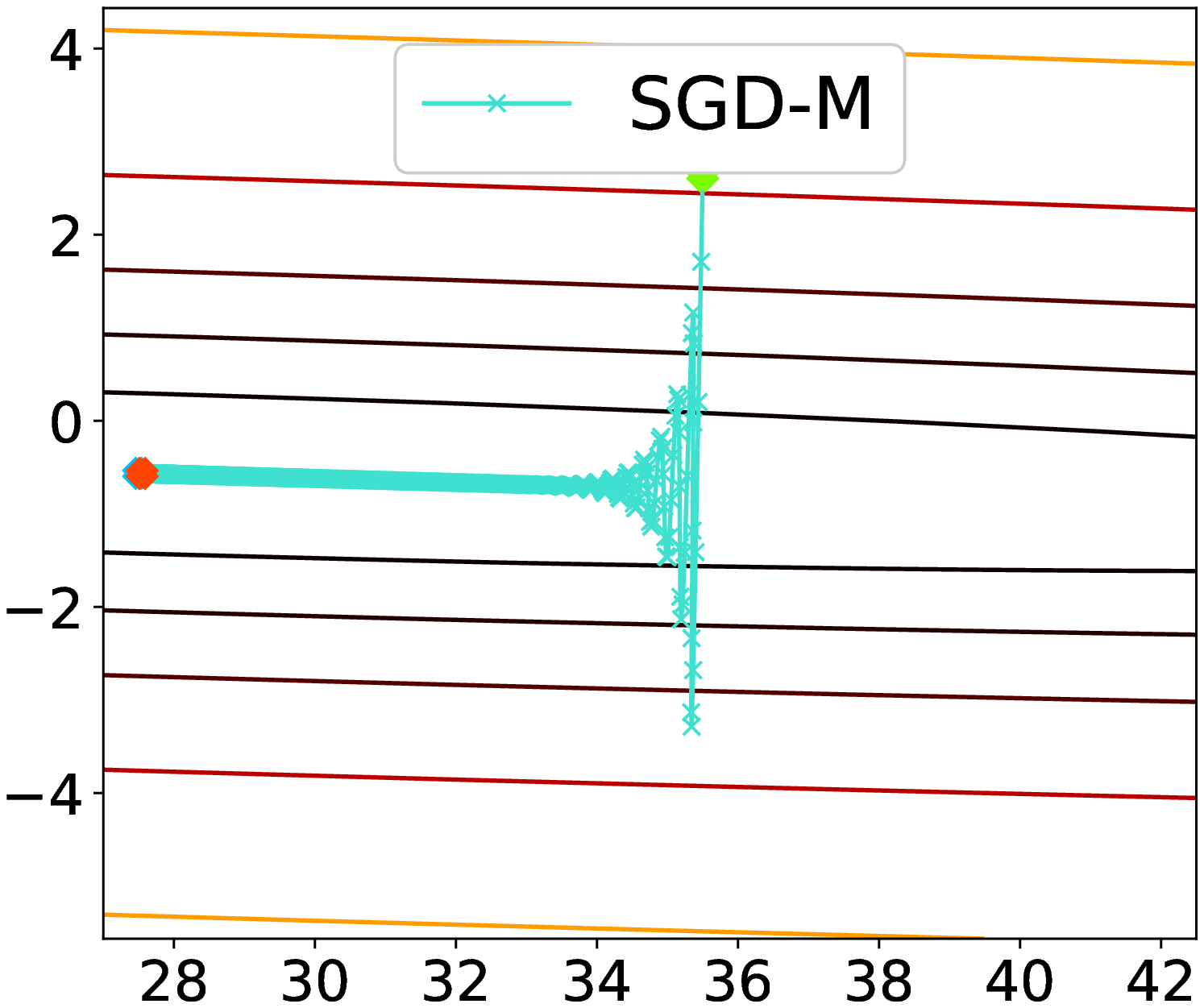}
\includegraphics[width=0.24\linewidth]{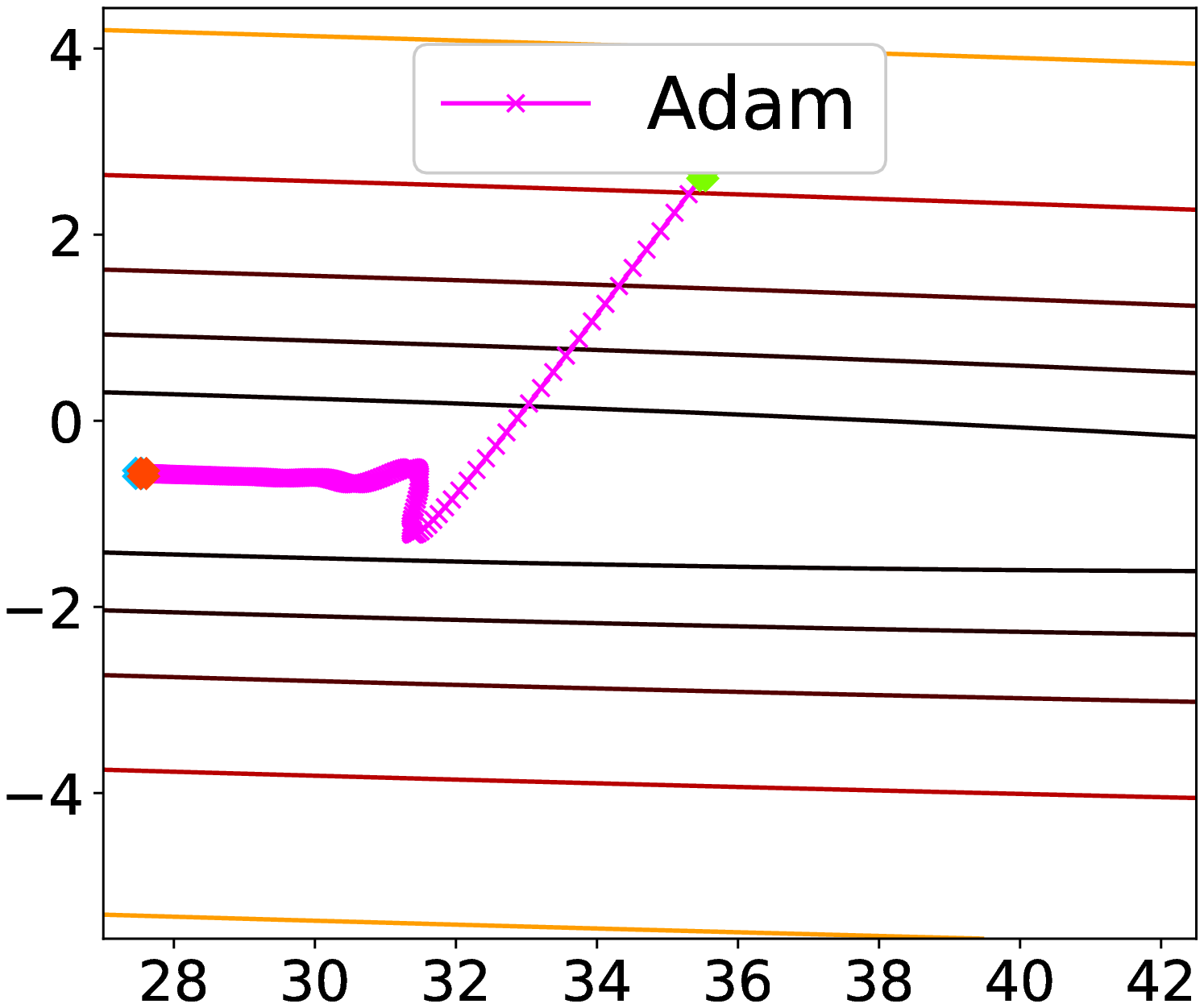}
\caption{Trajectory comparison of different optimisers on two quadratic optimisation problems (rows). Columns represent optimisers from left to right: \ourAcronym{}, SGD, SGD-M and Adam.  The point green point denotes the starting point and the orange point denotes the minima. \cut{Top Row: Mean of $Q_{0,0}$ and $Q_{1,1}$ are 0.3 and 14} 
Top row: Iterations to convergence are 23, 928, 104 and 45 respectively. 
Bottom Row: 
Iterations to convergence are 280, 27,328, 1,838 and 324 respectively. MetaMD is the fastest. }
\label{fig:meta_quadr_2d}
\end{figure}
}
\begin{figure}[t]
\centering
\includegraphics[width=0.24\linewidth]{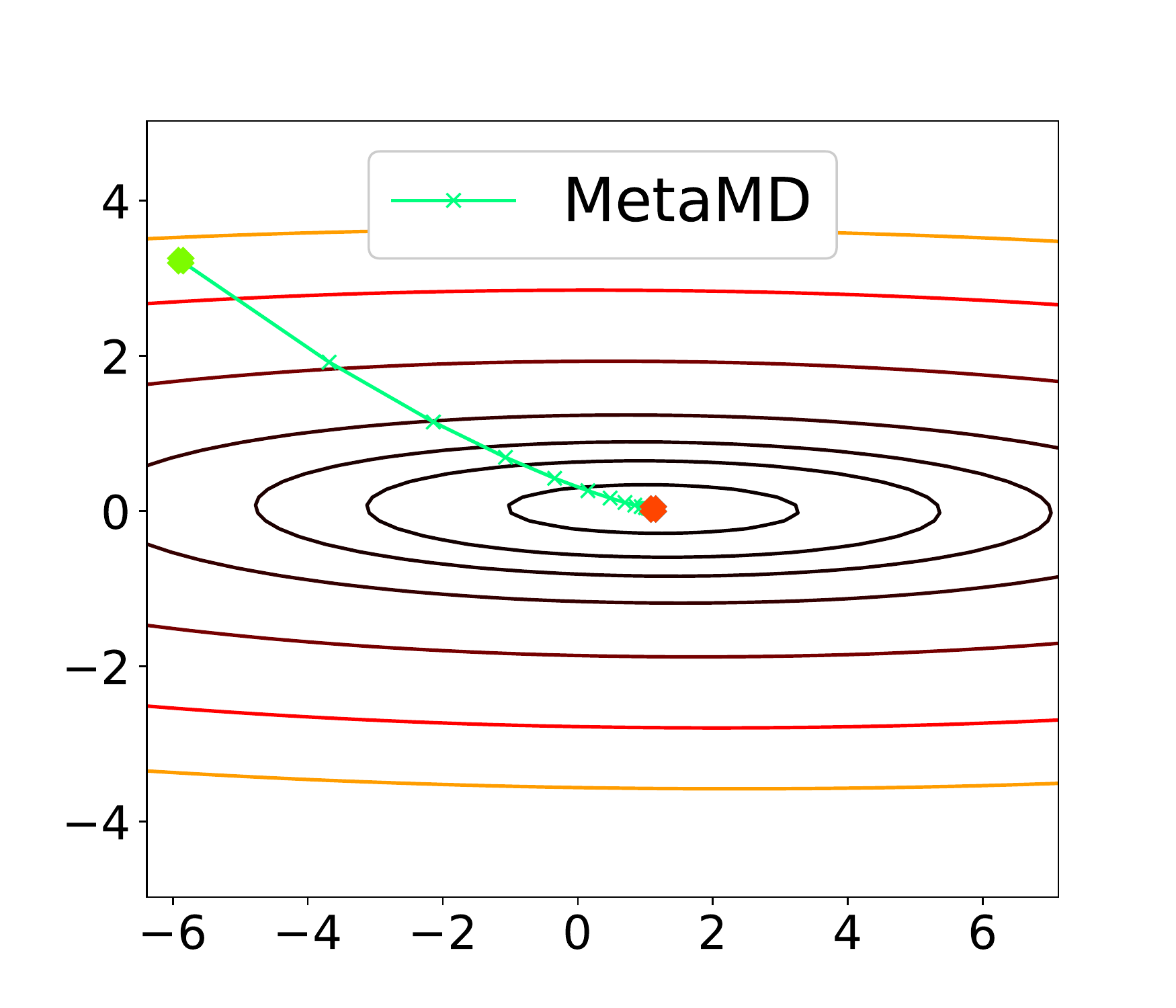}
\includegraphics[width=0.24\linewidth]{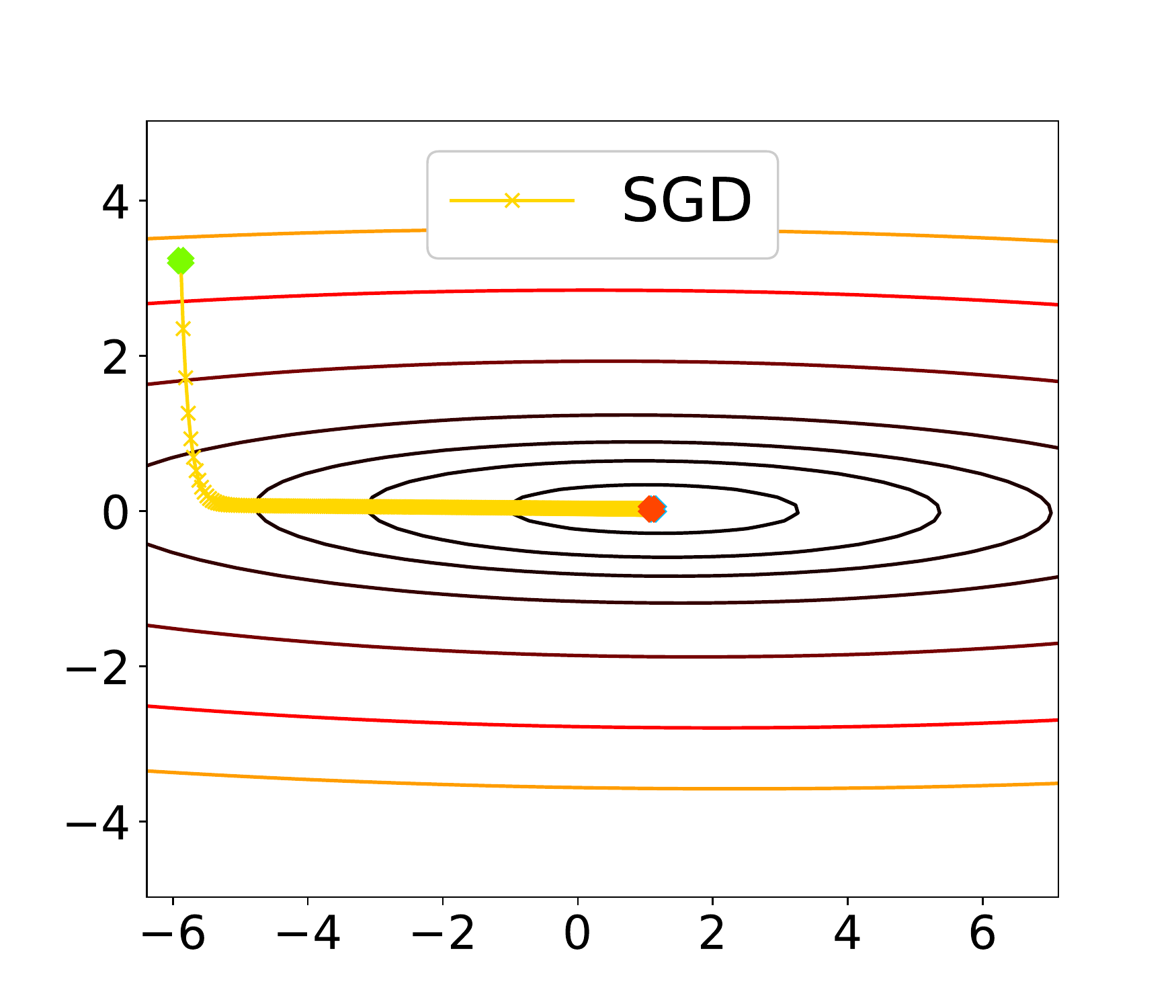}
\includegraphics[width=0.24\linewidth]{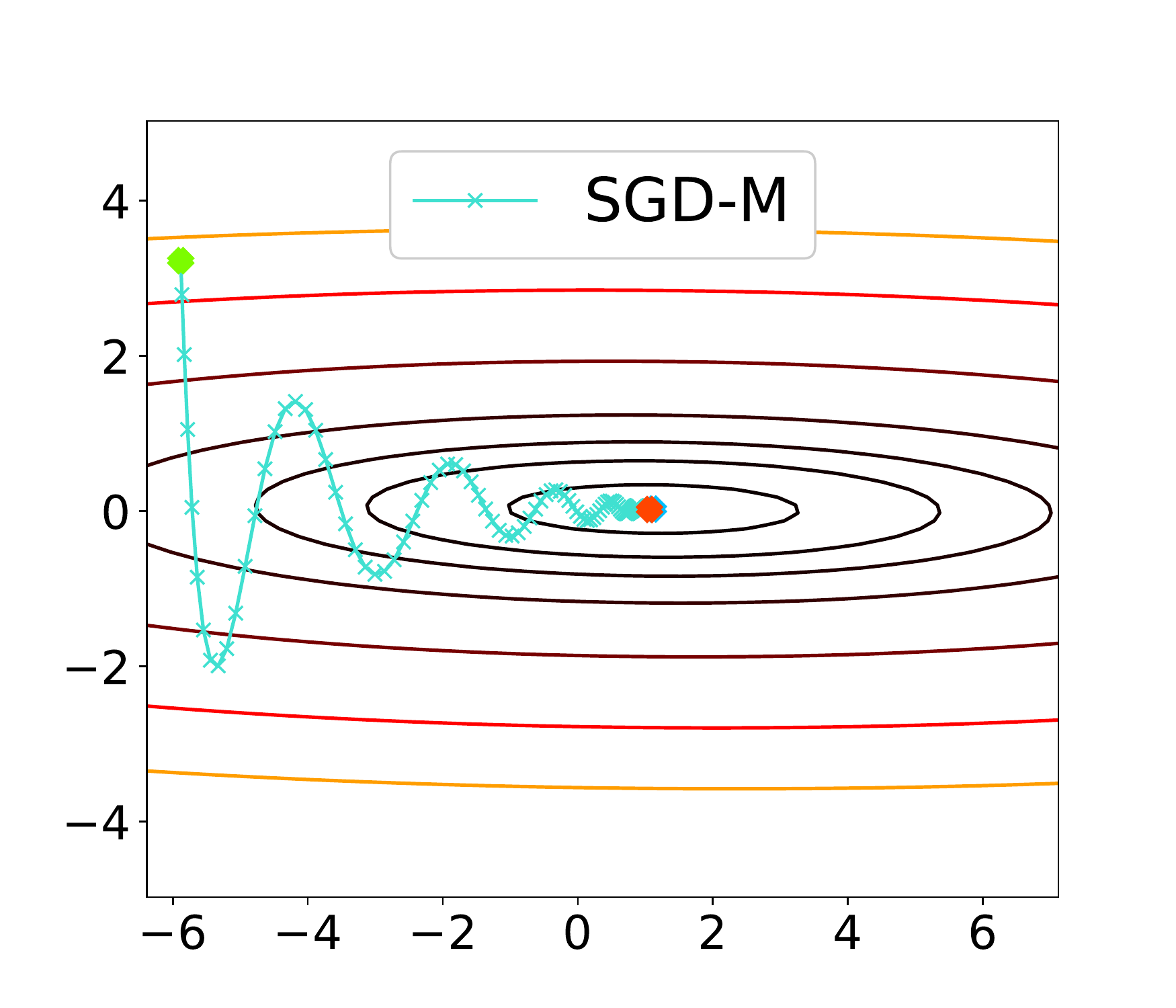}
\includegraphics[width=0.24\linewidth]{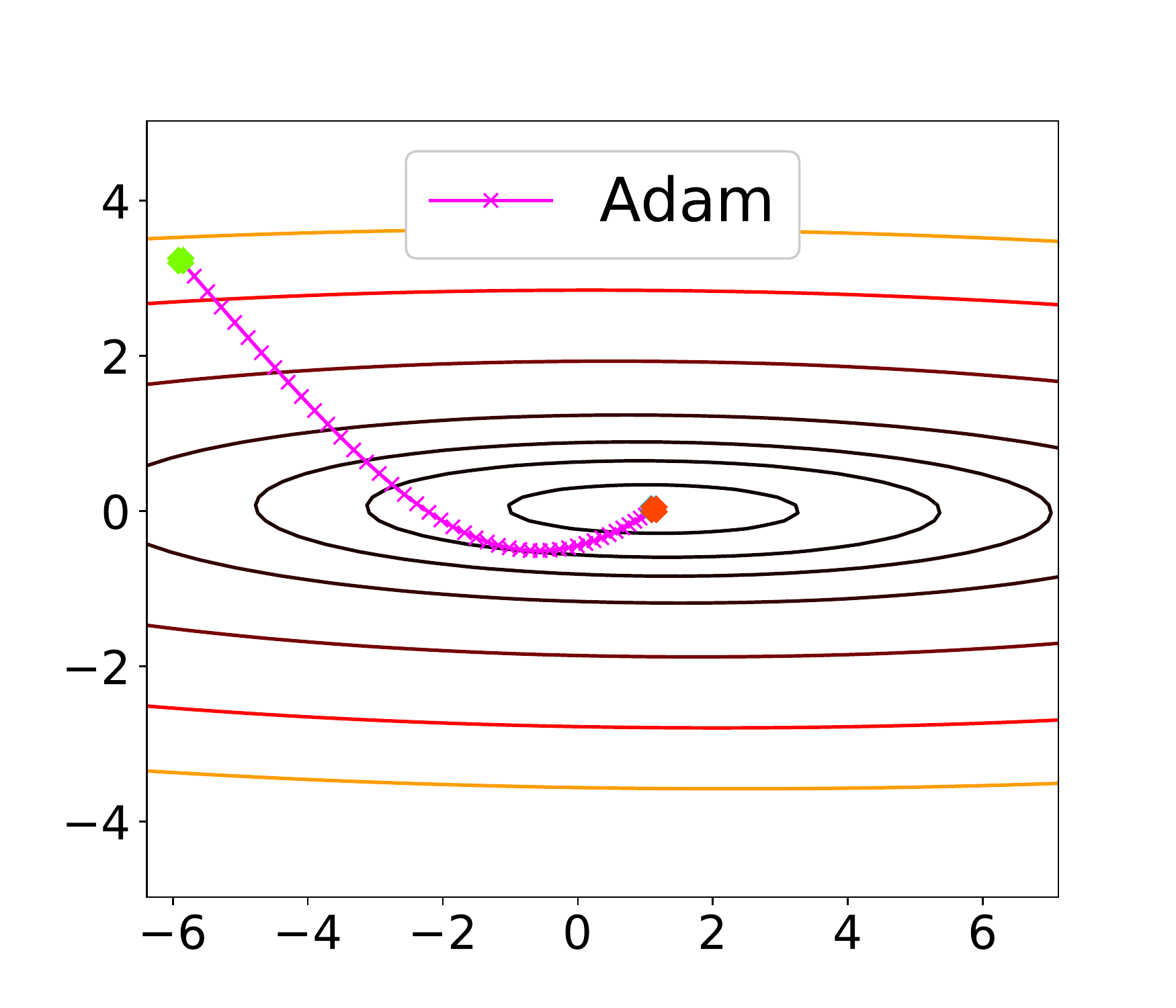}
\includegraphics[width=0.24\linewidth]{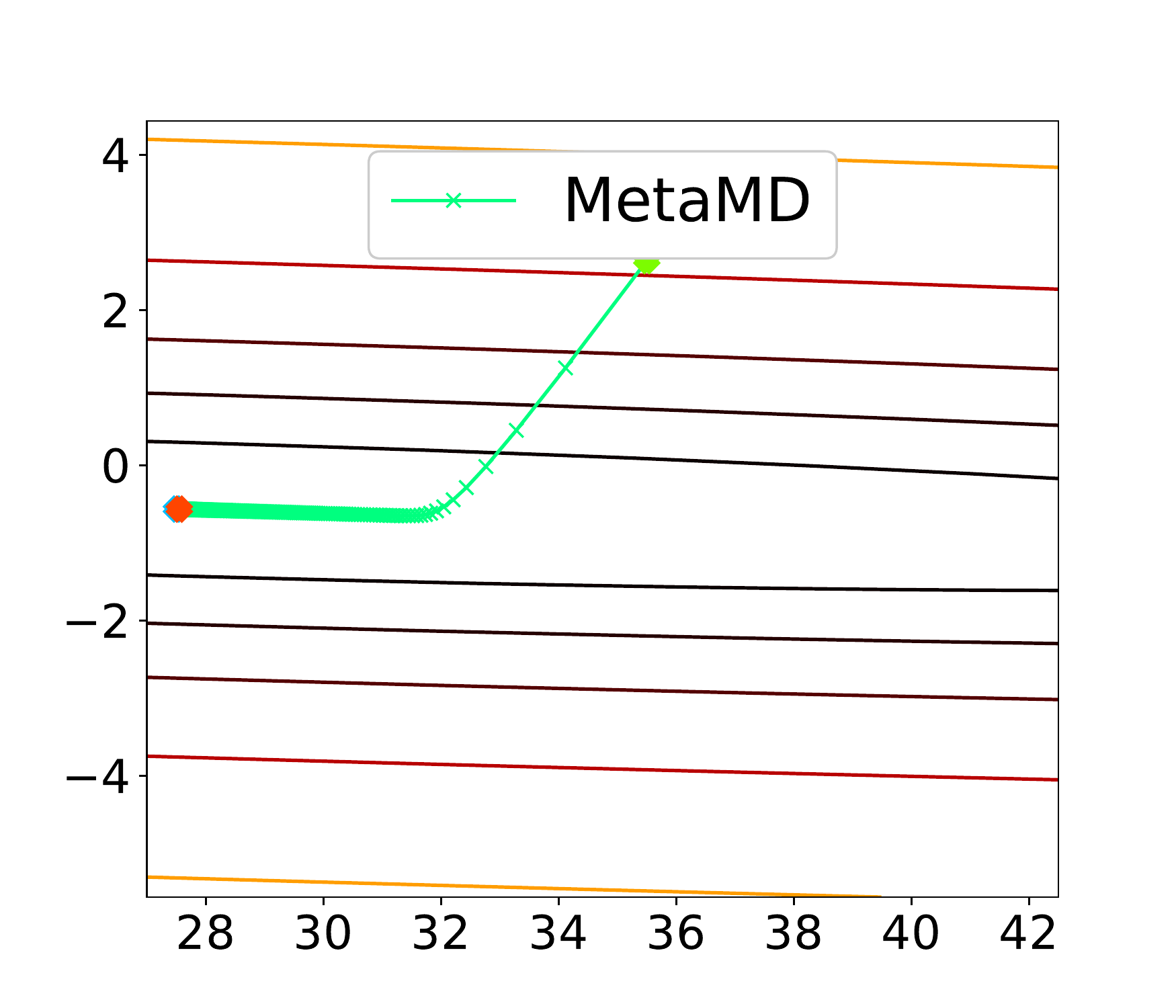}
\includegraphics[width=0.24\linewidth]{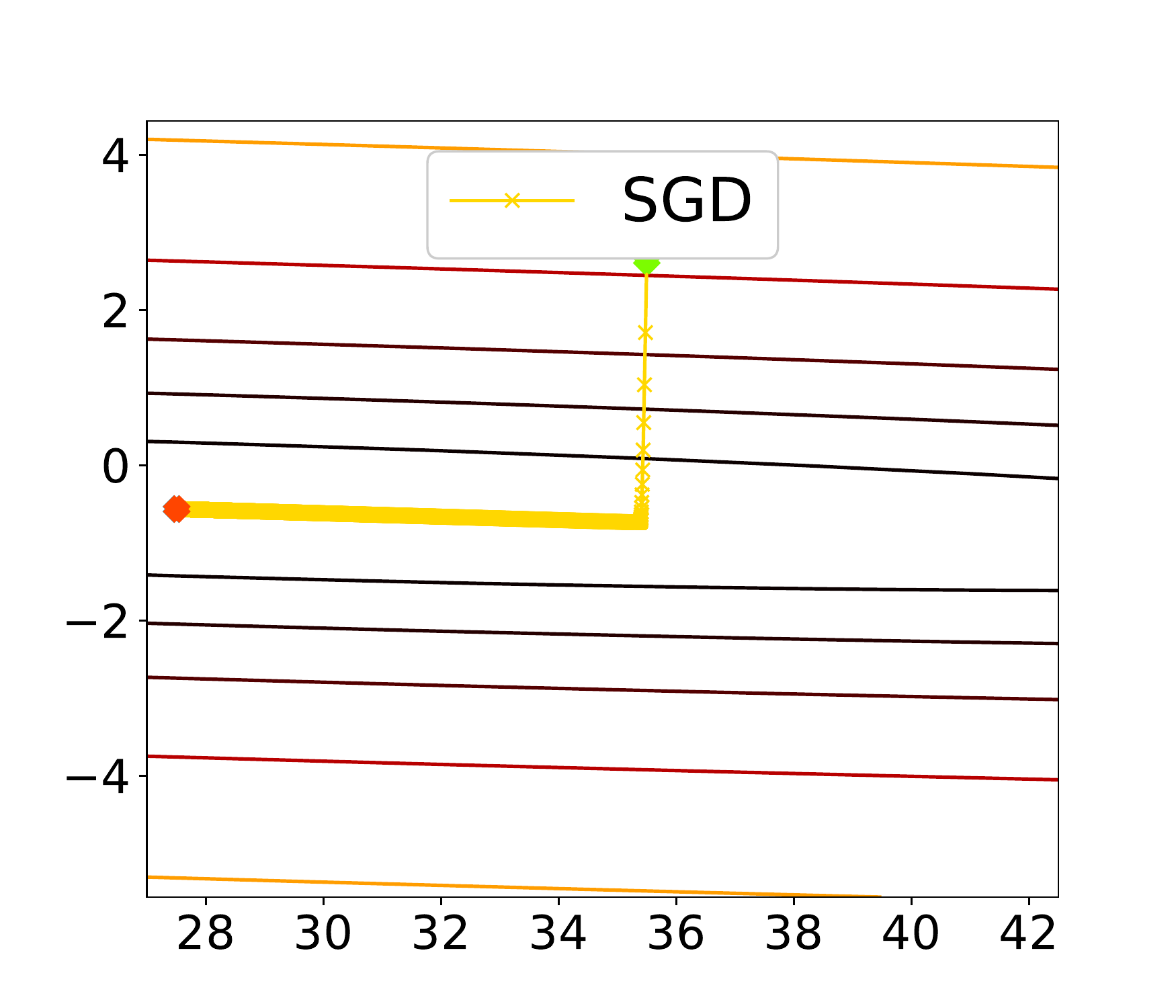}
\includegraphics[width=0.24\linewidth]{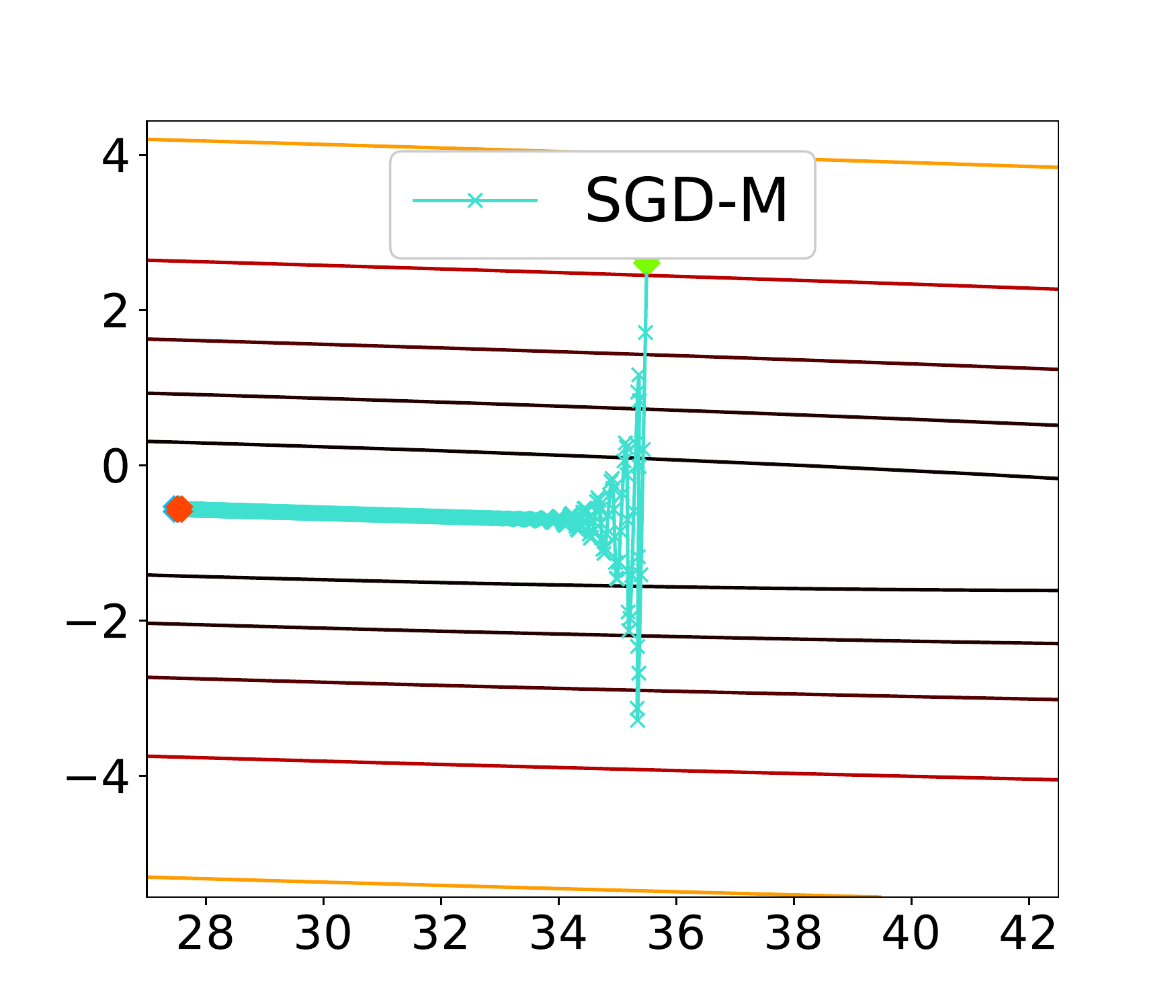}
\includegraphics[width=0.24\linewidth]{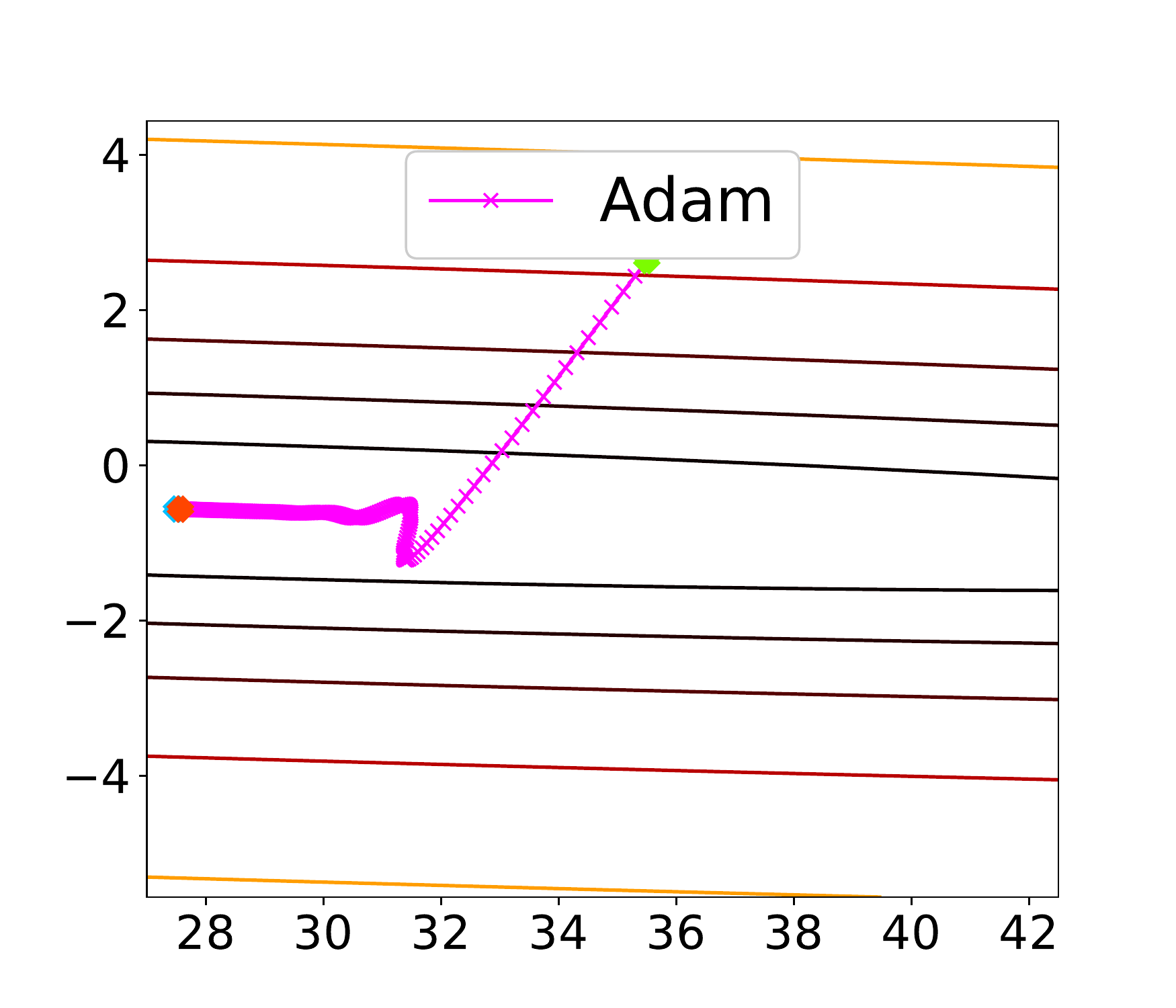}
\caption{Trajectory comparison of different optimisers on two quadratic optimisation problems (rows). Columns represent optimisers from left to right: \ourAcronym{}, SGD, SGD-M and Adam.  The point green point denotes the starting point and the orange point denotes the minima. \cut{Top Row: Mean of $Q_{0,0}$ and $Q_{1,1}$ are 0.3 and 14} 
Top row: Iterations to convergence are 23, 928, 104 and 45 respectively. 
Bottom Row: 
Iterations to convergence are 280, 27,328, 1,838 and 324 respectively. MetaMD is the fastest. }
\label{fig:meta_quadr_2d}
\end{figure}
\subsection{Synthetic Problem: Meta-Quadratic Optimisation}
\keypoint{Setup} We start the evaluation of \ourModel{} by creating a family of 2D quadratic optimisation problems from which we can sample a disjoint set of meta-training and meta-testing optimisation problems. We sample tasks of the form: 
\begin{align*}
 \min_\theta \theta^T Q \theta - b^T \theta 
\end{align*}
where $Q$ and $b$ are random variables. $b$ follows a Gaussian distribution with mean vector $[1,1]^T$ and identity covariance. To generate $Q$, we sample a two-dimension lower triangular matrix $C$ to construct the symmetric positive defined matrix $Q = C\cdot C^T$. We also illustrate problems with different loss flatness by specifying the mean of $Q_{0,0}$ and $Q_{1,1}$. 

\keypoint{Results} 
A comparison of optimisation trajectories on two kinds of meta-test quadratic problems is shown in Figure~\ref{fig:meta_quadr_2d}. All the optimisers are initialised in the same position and stopped when the norms of the gradient are smaller than the same  threshold. All the optimisers  reach the minima. But especially when varying the level of flatness, the competitors and even those with element-wise learning rates require much more iterations to converge compared with \ourAcronym{}.
\cut{
\begin{figure}[t]
\centering
\includegraphics[width=0.24\linewidth]{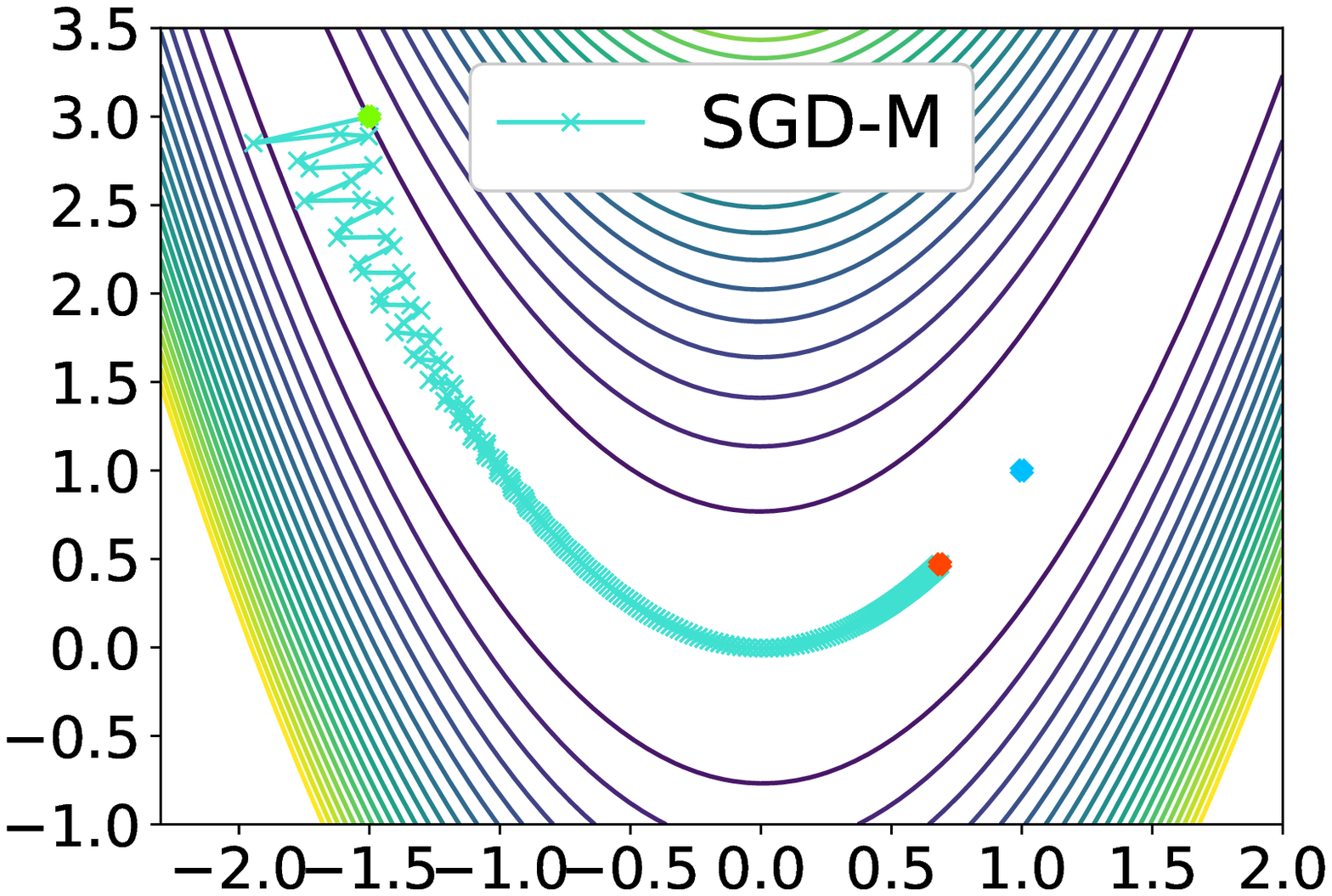}
\includegraphics[width=0.24\linewidth]{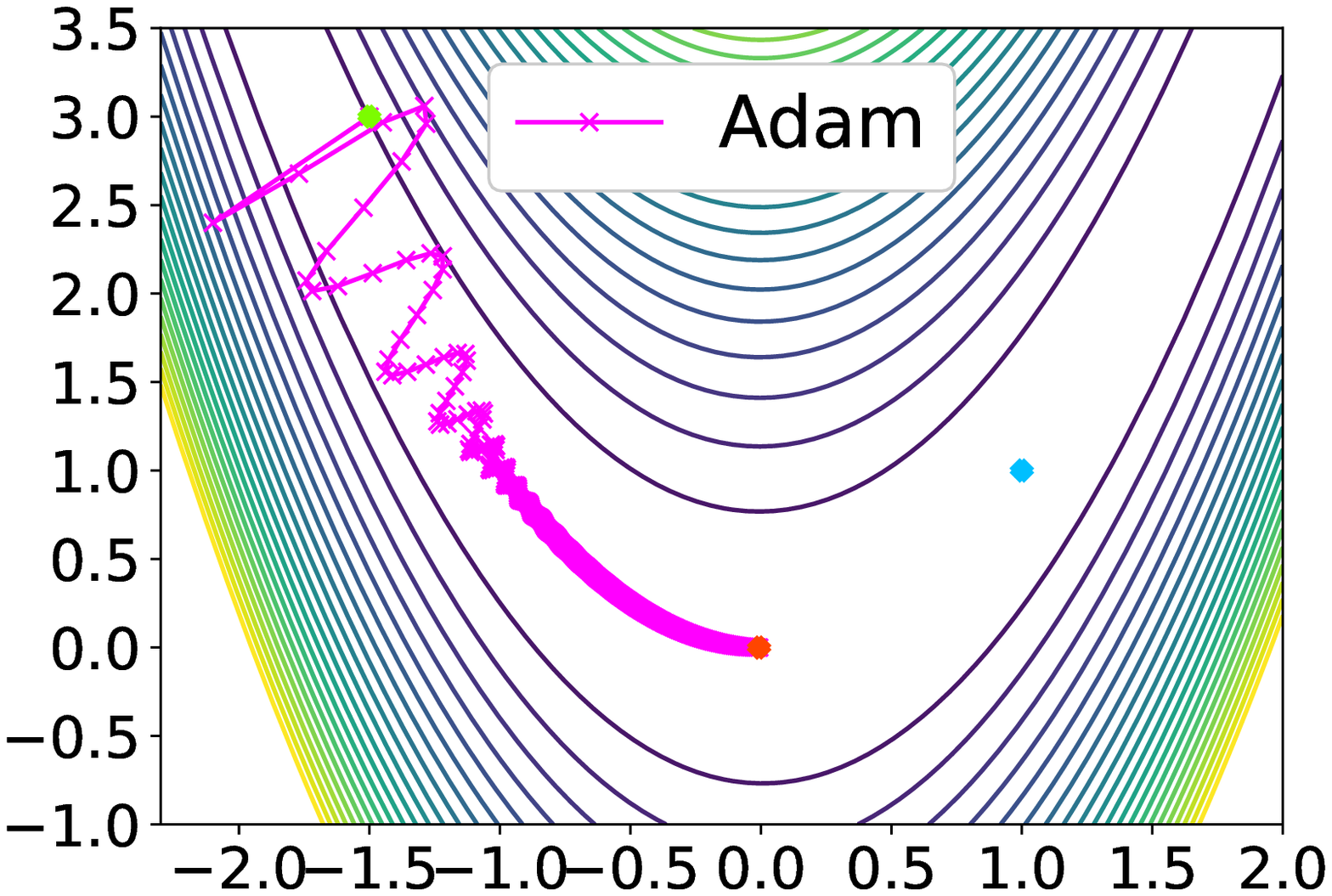}
\includegraphics[width=0.24\linewidth]{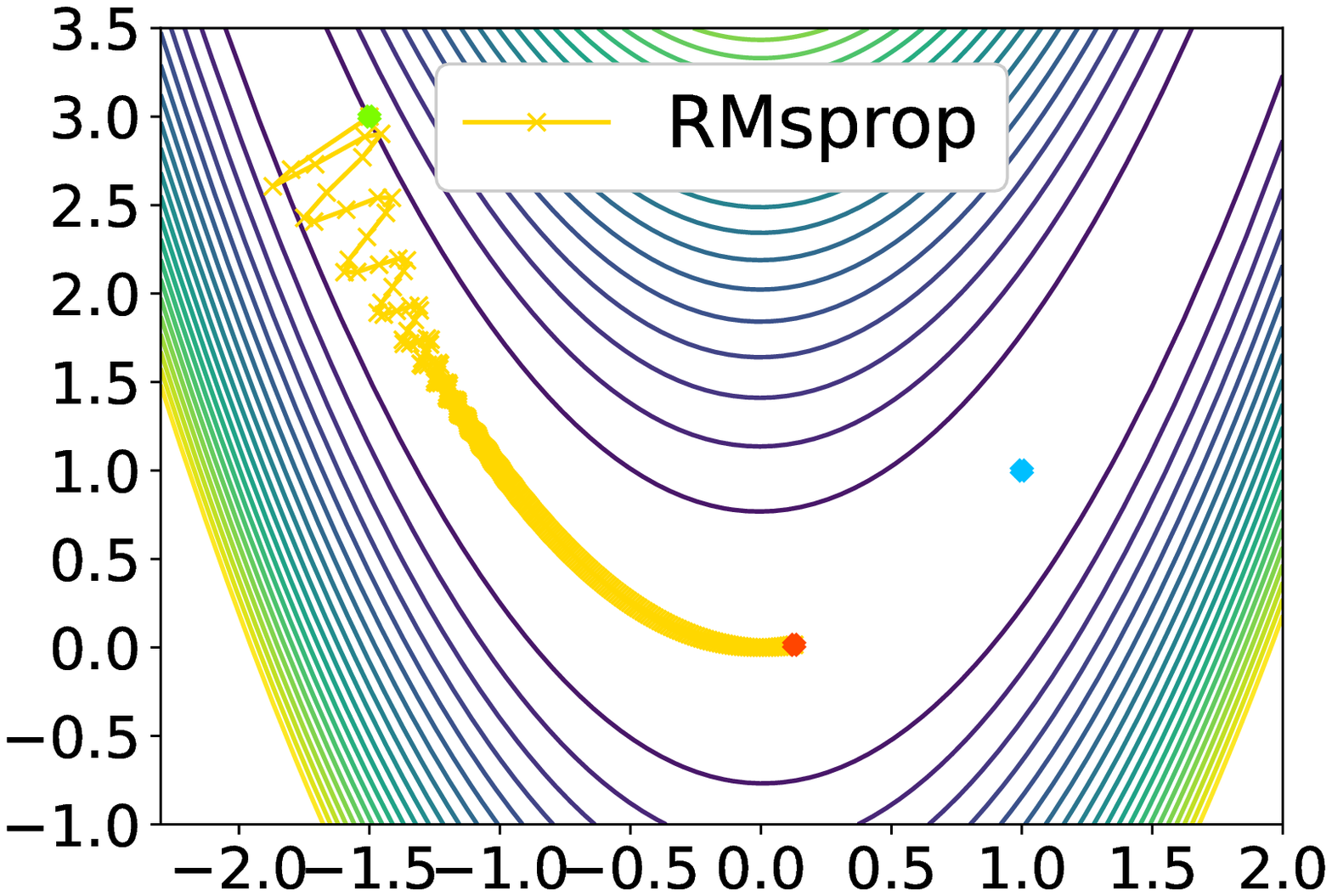}
\includegraphics[width=0.24\linewidth]{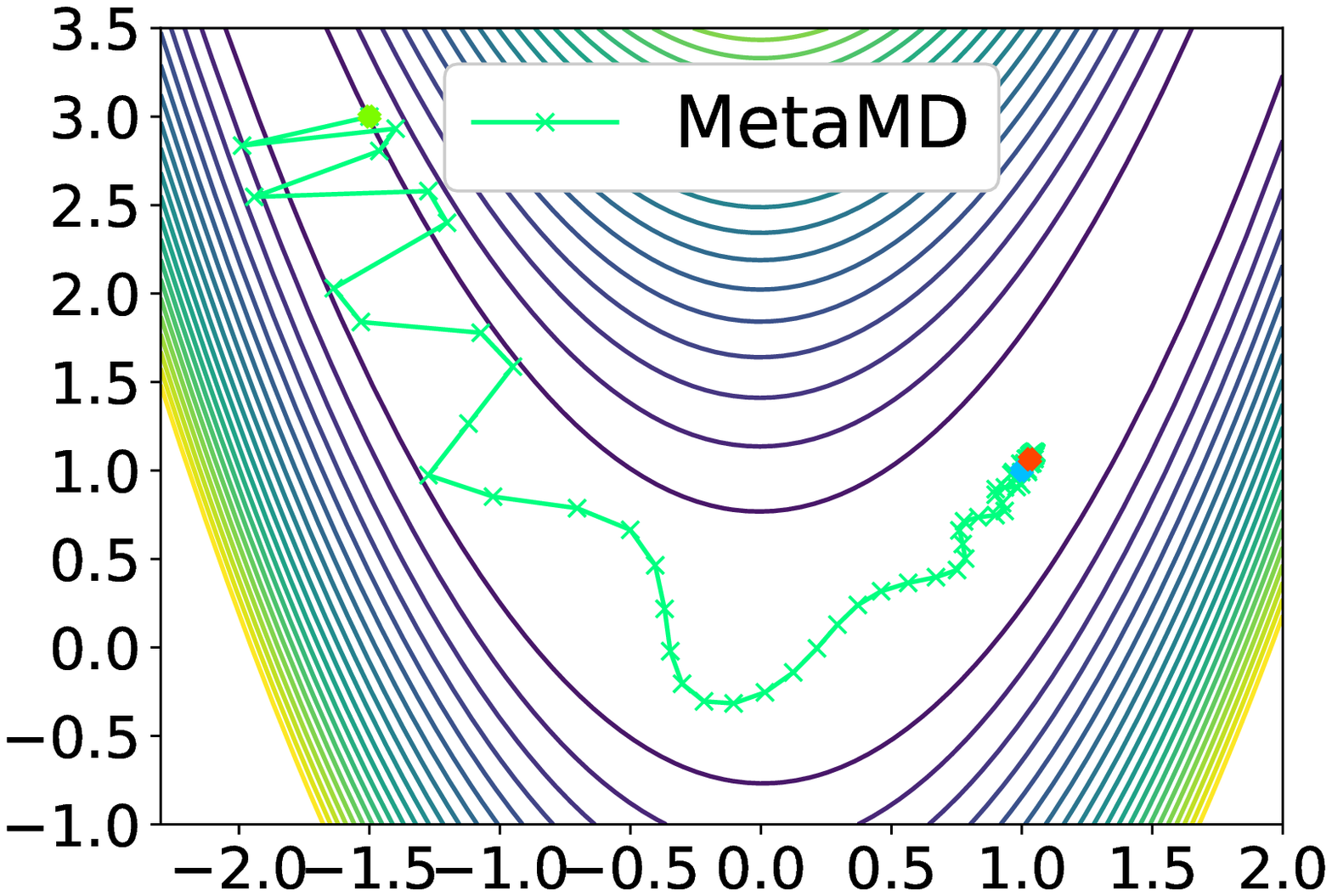}
\caption{Trajectory comparison of different optimisation algorithms on the quadratic optimisation problems. The green, orange and blue denote the starting points, the ending points, and minima respectively.}
\label{fig:rosenbrock_fuc}
\end{figure}

\subsection{Rosenbrock function}
Next we test our algorithm on the Rosebrock function, a non-convex function which is commonly used as a testbed for the optimisation algorithm. In this setting, we set the training iteration as 500, and carefully running tunning the learning for all the competiters, including Adam, SGD, SGD-M, RMSprop. The optimisation trajectories of each competiter and \ourModel{} are give in Figure~\ref{fig:rosenbrock_fuc}. Note that in this setting the meta-train and meta-test are both rosenbrock functions. In comparion, only \ourAcronym{} can reach the minima of Rosenbrock function, and the other competiters failed to reach the minima due the limited number of optimisation iterations. 
}
\subsection{Learning Mirror Descent for Neural Networks \label{sec:rotatedMNIST_MLP}}
\keypoint{RotatedMNIST and MLPs} We first evaluate optimiser learning for neural networks using the  RotatedMNIST dataset and a 3-layer MLP architecture. RotatedMNIST defines 6 domains by rotating the original MNIST dataset by 0, 15, 30, 45, 60 and 75 degrees. We use 5 domains for meta-training, and train \ourAcronym{} to convergence in the inner loop, and evaluate the performance on the held-out domain. This process is repeated, holding out each domain in turn as meta-test. The convergence curve is shown in Fig.~\ref{fig:rmnist_learning_curve}(left), and the testing performance in Table~\ref{tab:MLP_LeNet}(top). We can see that \ourAcronym{} converges rapidly and trains models with strong testing performance.  The hyperparameter tuning protocol for this and other experiments in this section is explained in Appendix~\ref{sec:hyper_tuning_shmodel}. 

\begin{figure}[t]
\centering
    \includegraphics[width=0.48\linewidth]{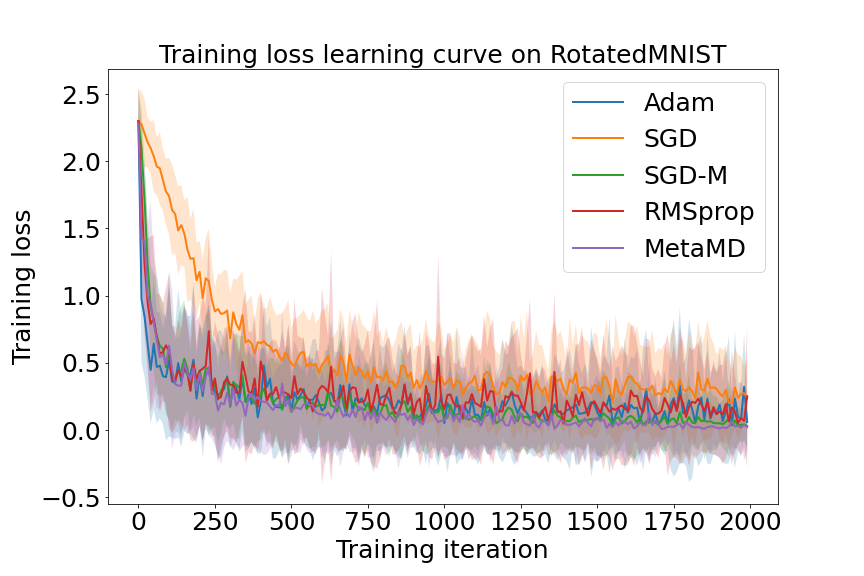}
    \includegraphics[width=0.48\linewidth ]{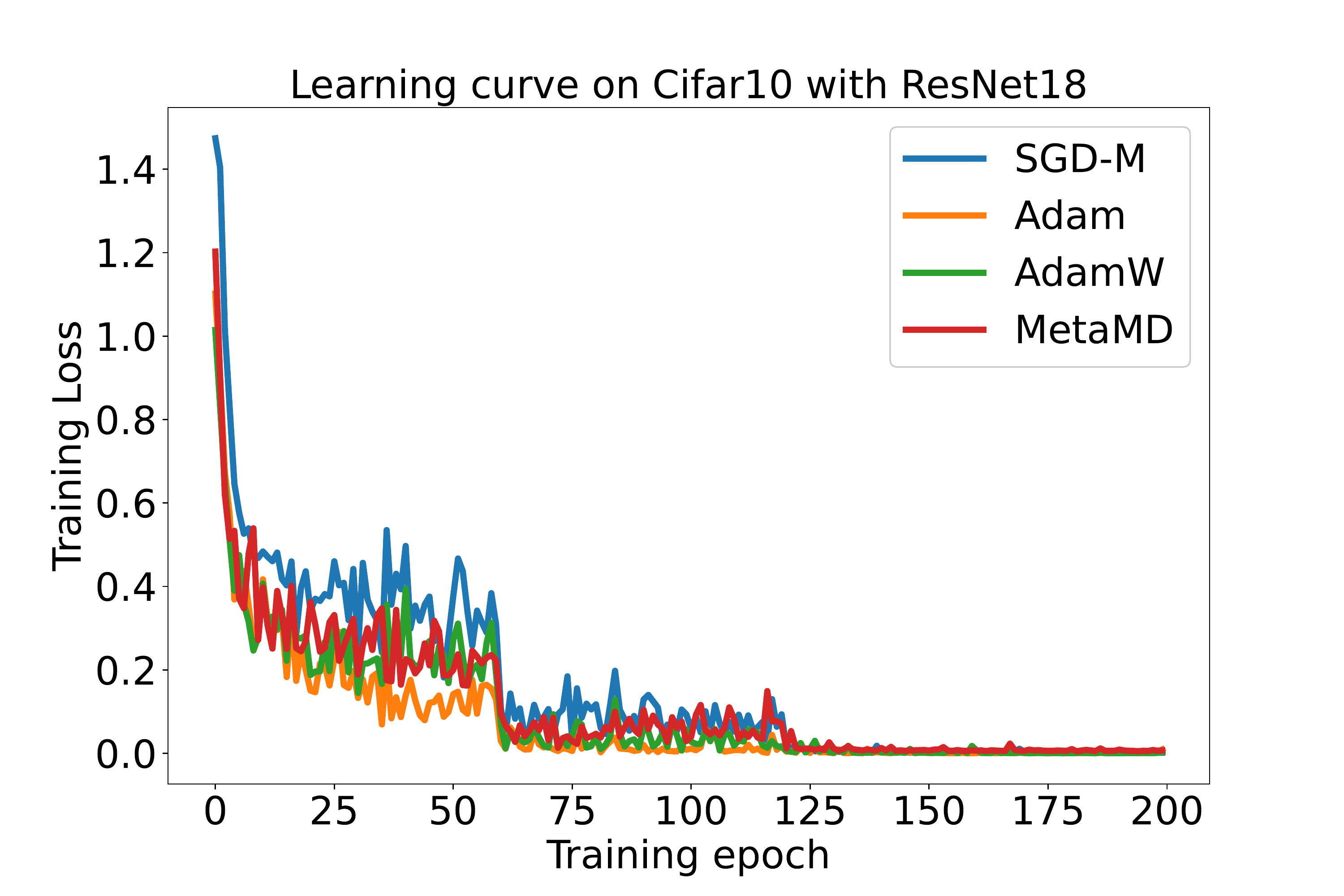}
\caption{Comparison of training loss convergence curves for different optimisers. Left: RotatedMNIST, averaged over all held-out domains. Right: Training loss curve of ResNet18 trained on CIFAR10, averaged over 3 trials.}
\label{fig:rmnist_learning_curve}
\end{figure}

\begin{table*}[t]
\centering
\caption{Test Accuracy (\%) on RotatedMNIST and DiverseDigits  with 3-layer MLP and LeNet respectivly. Each column is a test dataset, and MetaMD is trained on the other datasets.} \label{tab:MLP_LeNet}
\resizebox{0.9\textwidth}{!}{
\begin{tabular}{ l | l | c c c c c c}
\toprule
 &Test domain & 0 & 15 & 30 & 45 & 60 & 75 \\
\cmidrule{1-8}
\multirow{6}{*}{\rotatebox[origin=c]{90}{\textbf{3-Layer MLP}}}
&SGD               & 92.23 $\pm$ 0.57 & 91.91 $\pm$ 0.49 & 92.57 $\pm$ 0.32 & 92.89 $\pm$ 0.35 & 92.73 $\pm$ 0.32 & 92.36 $\pm$ 0.87 \\
&SGD-M             & 94.77 $\pm$ 0.58 & 94.64 $\pm$ 0.14 & 94.66 $\pm$ 0.29 & 94.67 $\pm$ 0.47 & 94.60 $\pm$ 0.47 & 94.47 $\pm$ 0.63 \\
&Adam              & 92.96 $\pm$ 0.58 & 93.29 $\pm$ 0.92 & 93.51 $\pm$ 0.84 & 93.69 $\pm$ 0.99 & 93.67 $\pm$ 0.35 & 92.98 $\pm$ 1.17 \\
&RMSprop           & 92.48 $\pm$ 0.49 & 93.56 $\pm$ 0.51 & 92.77 $\pm$ 0.50 & 93.58 $\pm$ 0.32 & 93.43 $\pm$ 0.32 & 93.14 $\pm$ 0.31 \\
\cmidrule{2-8}
&\ourAcronym{}     & \textbf{95.22} $\pm$ 0.70 & \textbf{95.18} $\pm$ 0.45 & \textbf{95.44} $\pm$ 0.27 & \textbf{95.34} $\pm$ 0.30 & \textbf{95.51} $\pm$ 0.57 & \textbf{95.12} $\pm$ 0.48 \\
\cmidrule{1-8}
& Test domain & MNIST & QMNIST & KMNIST & FashionMNIST & USPS & SVHN \\
\cmidrule{1-8}
\multirow{6}{*}{\rotatebox[origin=c]{90}{\textbf{LeNet}}}
&SGD            & 96.44 $\pm$ 0.91 & 96.23 $\pm$ 0.73 & 87.61 $\pm$ 1.87 & \textbf{88.95} $\pm$ 1.43 & 92.73 $\pm$ 1.13 & 85.44 $\pm$ 1.22 \\
&SGD+M          & 98.47 $\pm$ 0.16 & 97.21 $\pm$ 0.15 & 92.54 $\pm$ 0.62 & 86.44 $\pm$ 0.45 & 95.37 $\pm$ 0.24 & 86.26 $\pm$ 0.48  \\
&Adam           & 98.49 $\pm$ 0.17 & 98.10 $\pm$ 0.33 & 93.20 $\pm$ 0.82 & 87.36 $\pm$ 0.55 & 93.68 $\pm$ 0.38 & 87.07 $\pm$ 0.61  \\
&RMSprop        & \textbf{98.65} $\pm$ 0.21 & 98.30 $\pm$ 0.09 & 93.14 $\pm$ 0.87 & 87.45 $\pm$ 0.13 & 95.43 $\pm$ 1.06 & 87.01 $\pm$ 0.18  \\
\cmidrule{2-8}
&\ourAcronym{}  & 98.64 $\pm$ 0.12 & \textbf{98.41} $\pm$ 0.08 & \textbf{93.81} $\pm$ 0.33 & 87.72 $\pm$ 0.42 & \textbf{95.61} $\pm$ 0.70 & \textbf{87.59} $\pm$ 0.92   \\
\cmidrule{1-8}
\cmidrule{1-8}
\end{tabular}
}
\end{table*}


\keypoint{Diverse Digit Datasets and Small CNNs} 
Next we explore applying \ourAcronym{} to a more diverse set of datasets and CNN classifiers. A collect a group of datasets, which we denote as DiverseDigits, that includes: MNIST~\cite{lecun-mnisthandwrittendigit-2010}, QMNIST~\cite{qmnist-2019}, KMNIST~\cite{clanuwat2018deep}, FashionMNIST~\cite{xiao2017fashion}, USPS~\cite{hull1994database} and SVHN~\cite{netzer2011reading}. We train LeNet classifier using \ourAcronym{}, resizing all images to $28\times28$ greyscale. The same leave-one-dataset-out protocol is used: Each dataset is held out in turn for evaluation after MetaMD is trained on on the other datasets. Compared to the previous RotatedMNIST experiment, the distribution of tasks used for meta-training and meta-testing is now more diverse and challenging. Due to the greater cost of training the base model here, we use $T=500$ iterations for the inner loop, and leave efficient meta-learning under longer-horizons as future work. We compare all methods fairly by a common hyperparameter (learning rate, weight-decay, etc) tuninng protocol for meta-test. Specifically, we perform BayesOpt with respect to meta-test validation accuracy with 25 iterations for each competitor and more detail is given in Appendix~\ref{sec:hyper_tuning_shmodel}. 

The results averaged over 3 meta-test trials are shown as testing performance at convergence in Table~\ref{tab:MLP_LeNet}(bottom) and selected meta-test learning curves in Fig.~\ref{fig:mnist_lenet}, with the remaining learning curves given in Appendix~\ref{sec:whole_lenet_mnist}. We can see that \ourAcronym{} is clearly faster than SGD and SGD-M in training convergence (Fig.~\ref{fig:mnist_lenet}), while typically producing models with the strongest generalisation error (Table~\ref{tab:MLP_LeNet}). It is noteworthy that MetaMD exhibits strong cross-dataset generalisation here, corroborating our Theorem~\ref{thm:generalisation} on cross-task optimiser generalisation. 


\begin{figure}[t]
    \centering
    \includegraphics[width=0.3\linewidth]{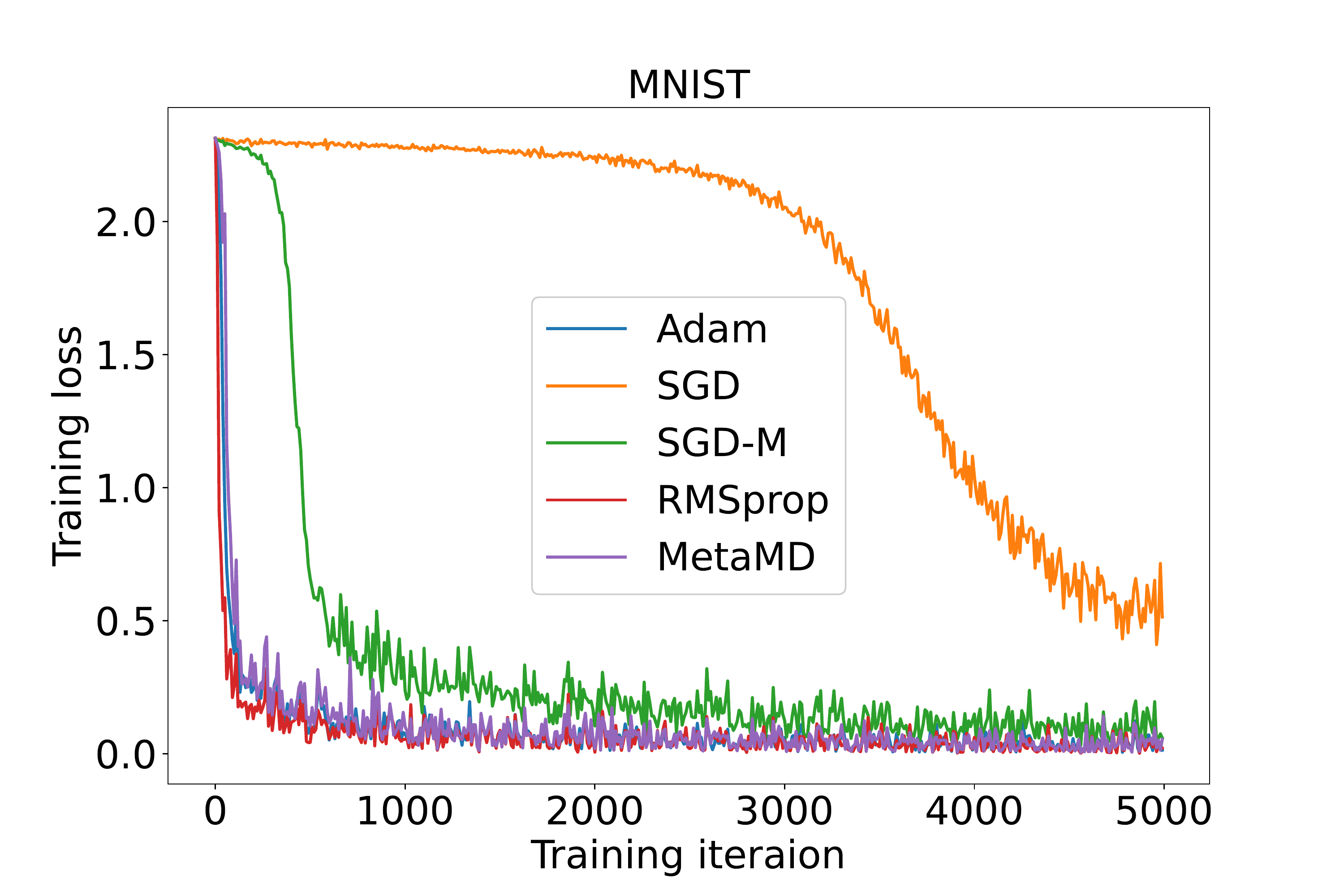}
    \includegraphics[width=0.3\linewidth]{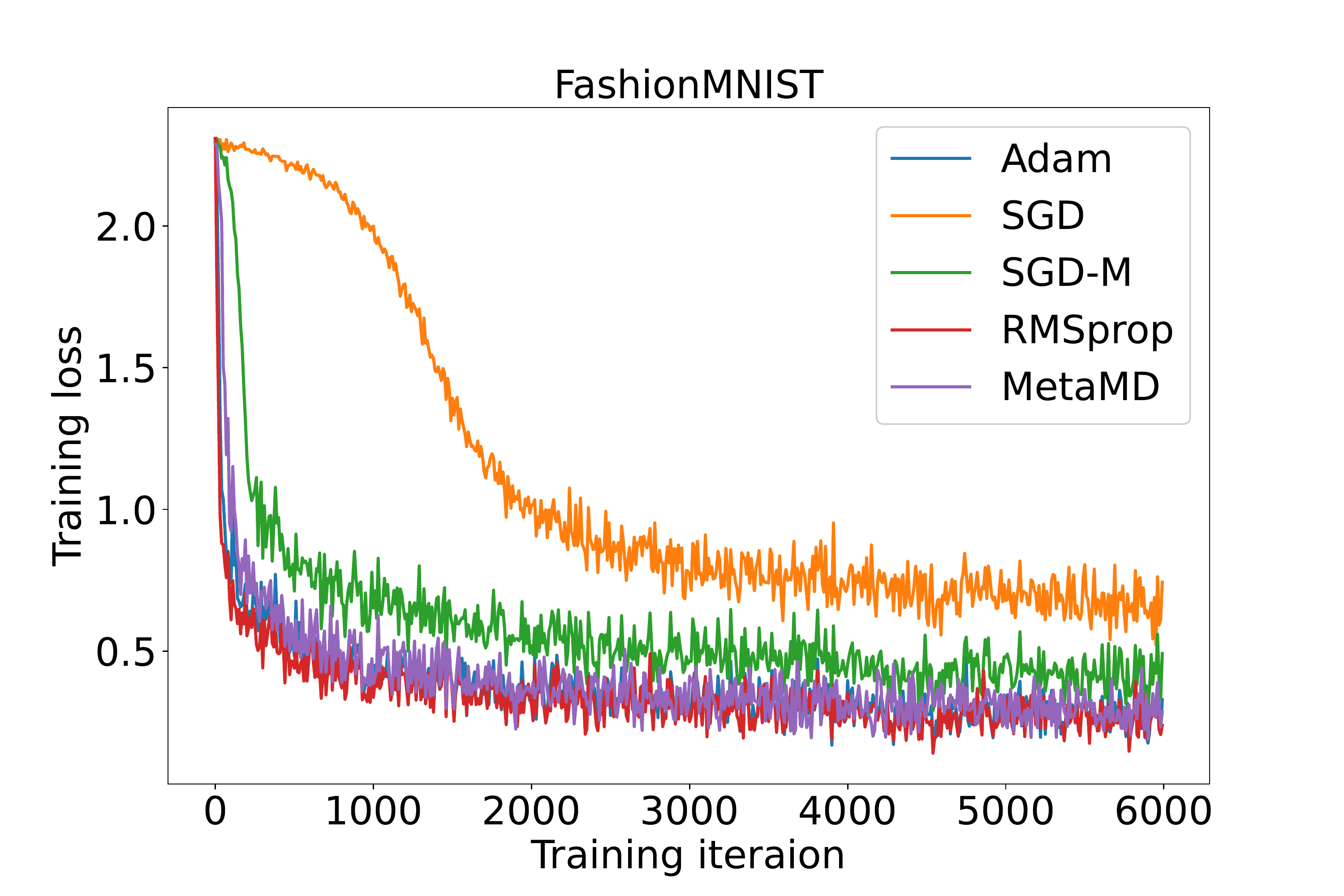}
    \includegraphics[width=0.3\linewidth]{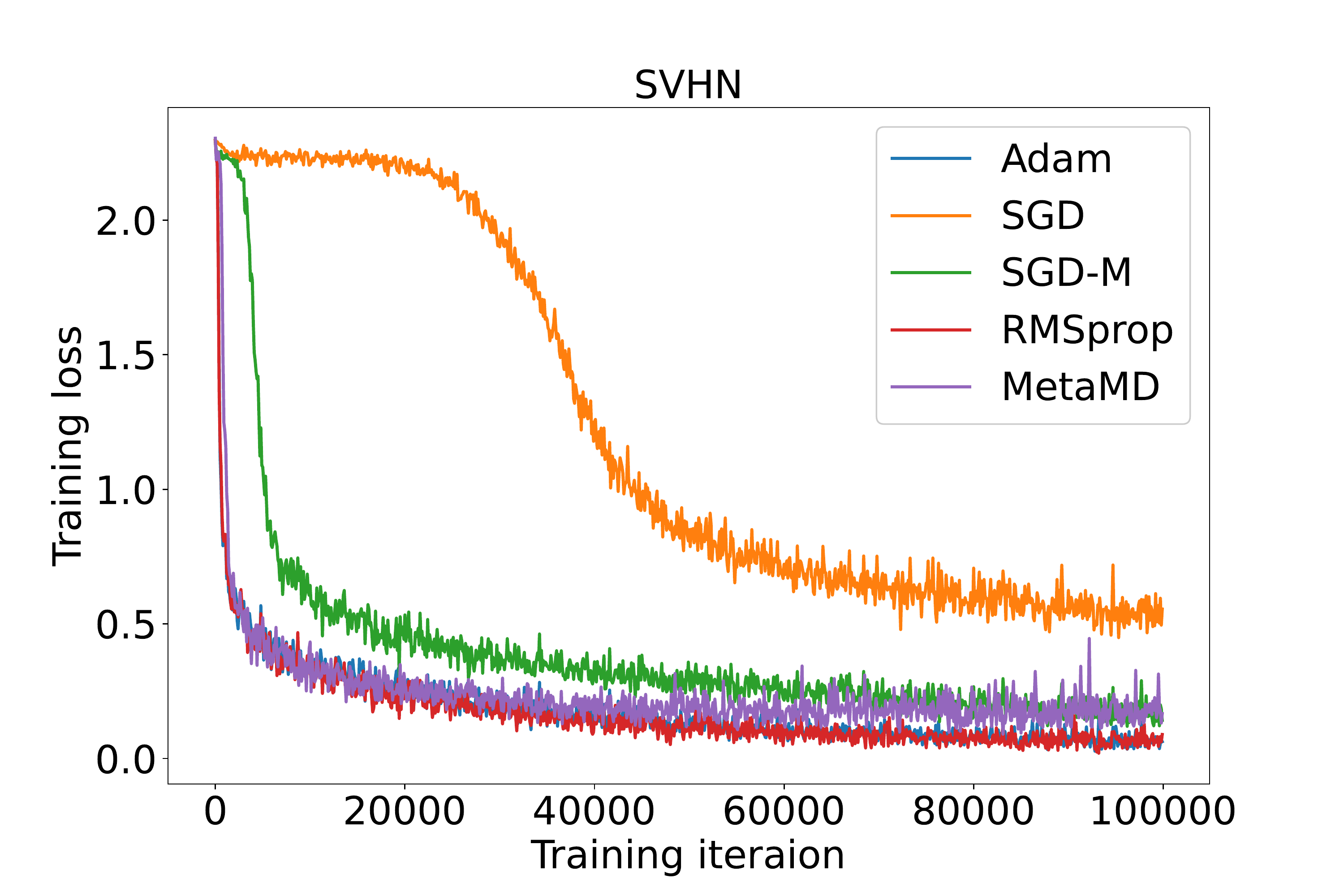}
\caption{Learning curves for DiverseDigits. From left to right: MNIST, FashionMNIST and SVHN.}
\label{fig:mnist_lenet}
\end{figure}
\begin{wrapfigure}{r}{0.4\linewidth}
\begin{minipage}[H]{\linewidth}
\resizebox{\linewidth}{!}{
\begin{tabular}{ l | c c c c}
    \toprule
    Method & SGD-M & Adam & AdamW  & MetaMD\\
    \midrule
    Accuracy & 91.43  & 91.29 & 92.64 & 93.74  \\
    \bottomrule
    \end{tabular}
    }
    \caption{Test accuracy on CIFAR10 using Resnet18.} 
   \label{tab:resnet18_cifar10}
\end{minipage}
\end{wrapfigure}
\keypoint{Application to ResNet18 and CIFAR10} We finally focus on training the deeper and larger ResNet18 on CIFAR10 as a held out testing task. To this end we construct a suite of meta-training datasets by combining STL10~\citep{coates2011analysis} and DiverseDigits from the previous setting. ResNet18+CIFAR10 is a well-studied problem with lots of known tuning tricks for standard optimisers. For fair comparison, we therefore tune all methods with exactly the same BayesOpt-based hyperparameter tuning protocol, based on CIFAR10 validation performance. 

The results in terms of learning curves and testing accuracy at convergence are shown in  Figure~\ref{fig:mnist_lenet}(right) and  Table~\ref{tab:resnet18_cifar10}, averaged over three complete meta-test trials. MetaMD is faster than standard SGD, while providing better test accuracy than both SGD and Adam.  \begin{wrapfigure}{R}{0.4\linewidth}
  \begin{center}
    \includegraphics[width=\linewidth]{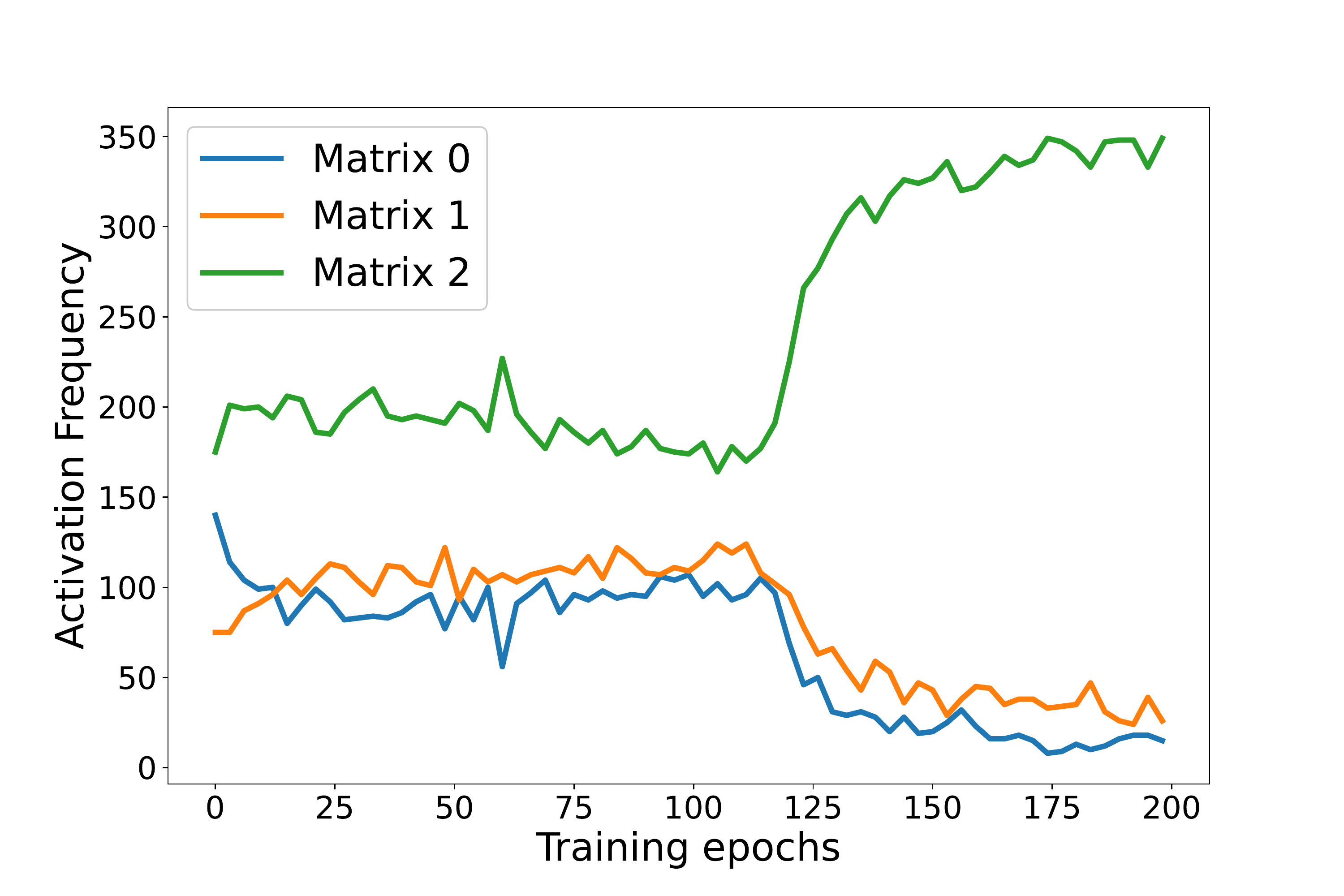}
  \end{center}
   \caption{Activation Frequency of each matrix during training.}
   \label{fig:activation_trend_plot}
\end{wrapfigure}


\keypoint{Further Analysis}
A technical contribution in this work is to propose mixture-of-mahanobis distances as an expressive yet efficiently computable optimiser parametrisation. This is in contrast  to many existing meta-learned optimisers  \citep{li2017meta,antoniou2018maml}, which learn a single set of learning rates. To analyse this, we report the activation frequency of each mahalanobis distance over training epochs of the base model. We can see that the learned dynamics tend to prefer one matrix only slightly at first, and then more substantially after the first 100 epochs. This demonstrates that MetaMD makes use of this additional degree of freedom compared to standard optimizers.  

\cut{
\begin{figure}[tb]
\centering
    \begin{minipage}{.48\linewidth}
    \resizebox{0.9\linewidth}{!}{
    \includegraphics[width=\linewidth]{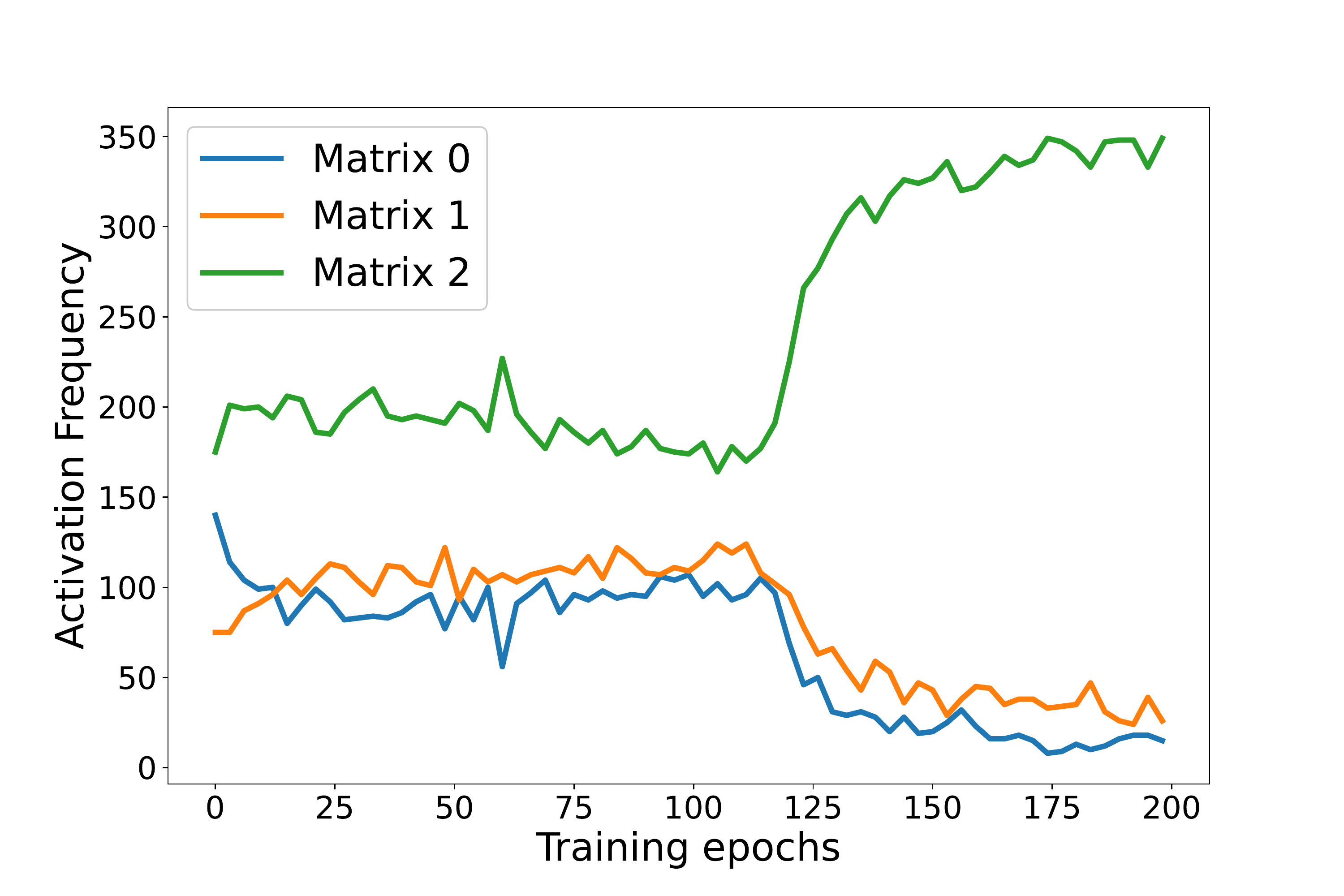}
    }
    \caption{Activation Frequency of each matrix during training}
    \label{fig:activation_trend_plot}
    \end{minipage} 
    \hfill
    \begin{minipage}{.48\linewidth}
    \begin{table}[H]
    \resizebox{\linewidth}{!}{
    \begin{tabular}{ l | c c c c}
    \toprule
    Matrix & 0 & 1 & 2  & total\\
    \midrule
    Accuracy & 93.48  & 93.56 & 93.28 & 93.74  \\
    \bottomrule
    \end{tabular}
    }
    \caption{Test Accuracy of \ourAcronym{} with applying only one learned Mahalnobis distance matrix} 
    \label{fig:single_matrix_performance}
    \end{table}
    \end{minipage} 
\end{figure}
}

\cut{
\begin{figure}[tb]
    \centering
    \begin{minipage}{0.49\linewidth}
    \resizebox{0.9\linewidth}{!}{
    \includegraphics[width=\linewidth, height=4cm]{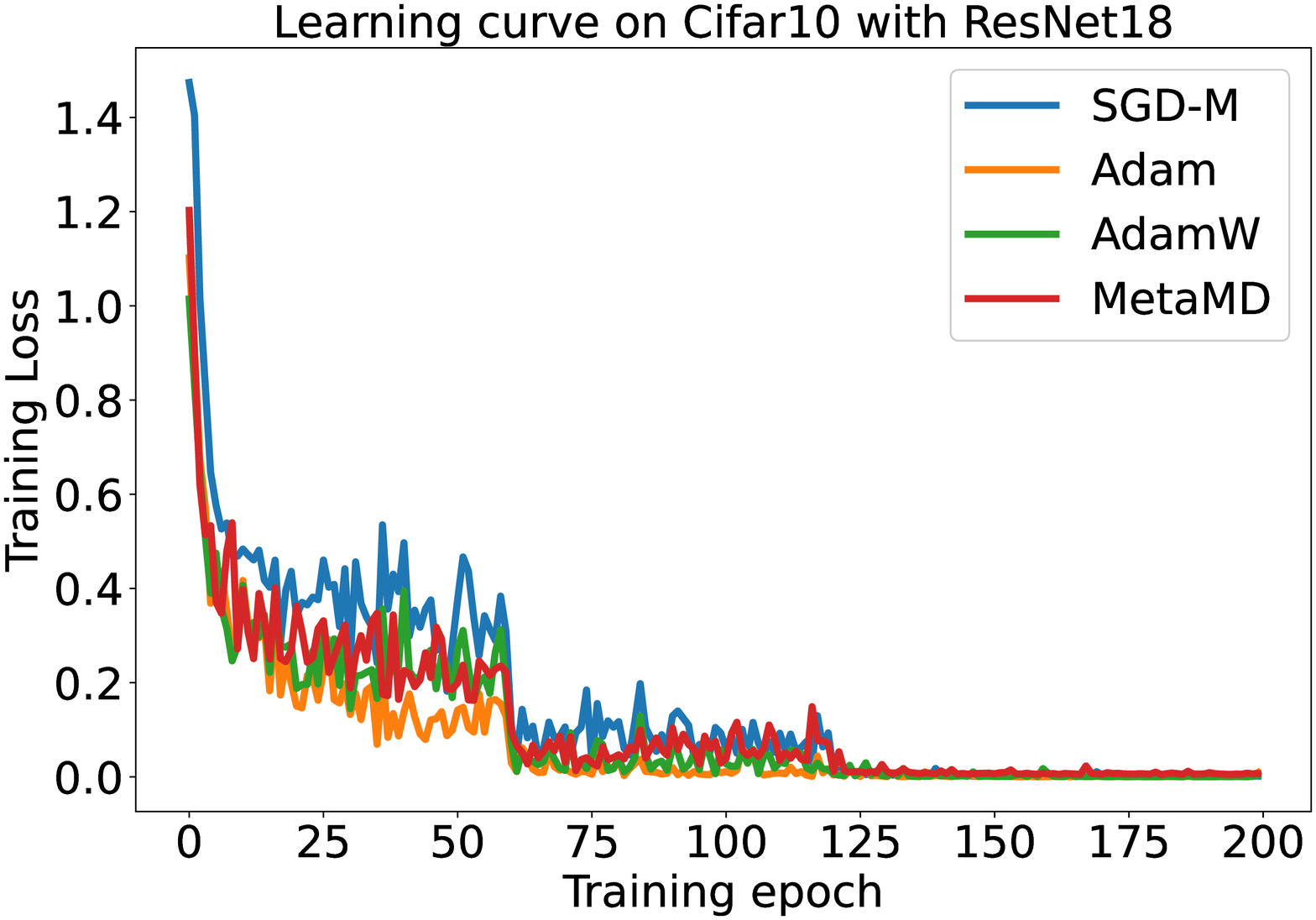}
    }
    \caption{Learning curve comparison of different optimiser on Cifar10 with ResNet18.}
    \label{fig:cifar10_resnet18_converge}
    \end{minipage} \hfill
\begin{minipage}{0.45\linewidth}
\begin{table}[H]
\begin{center}
\resizebox{\linewidth}{!}{
\begin{tabular}{ l | c c c c}
\toprule
Method & SGD-M & Adam & AdamW  & \ourAcronym{}\\
\midrule
Accuracy & 91.43  & 91.29 & 92.64  & \textbf{ 93.74}  \\
\bottomrule
\end{tabular}
}
\end{center}
\caption{Test performance on the ResNet18 and Cifar10 setting with different optimisers} 
\label{tab:resnet18_cifar10}
\end{table}
\end{minipage} 
\label{fig:resnet18_cifar10}
\end{figure}
}

\section{Conclusion}
We explored meta-learning optimisers from the Mirror Descent perspective. More precisely, an algorithm is proposed to meta-learn a Bregman Divergence to manipulate the gradient for updating the base model. With an efficient instantiation based on mahalanobis distances, this can be interpreted as a mixture of elementwise learning rates. Our approach has clear theoretical motivation by optimizing a regret bound on the convergence rate, and has both a convergence guarantee and a cross-dataset generalisation guarantee. Empirically, our results demonstrate rapid convergence compared to SGD and strong generalisation vs other fast optimisers such as Adam.

An obvious limitation of our empirical results is that we have compared to other fast hand engineered optimisers, but not to other meta-learned optimisers. In future work we will compare to alternative meta-learned optimisers and continue to search for better  paramaterisations of our Bregman divergence $B_\phi$. 


\clearpage

\bibliography{6_reference}
\bibliographystyle{iclr2022_conference}

\newpage
\appendix
\section{Appendix}
\subsection{Derive of the closed form mirror loop \label{closed_inverse_mapping}}
In our setting, the mirror loop is described as:
\begin{align*}
\theta_{t+1} = \argmin_{\theta} \langle \nabla_{\theta} \mathcal{L}(\theta_t), \theta \rangle + \frac{1}{2 \eta} B_{\phi}(\theta || \theta_t) 
\end{align*}
for simplicity but keeping the generalisity we rearrange the equation as:
\begin{align*}
\theta_{t+1} = \argmin_{\theta} \eta \langle \nabla_{\theta} \mathcal{L}_{tr}(\theta_t), \theta \rangle +  B_{\phi}(\theta || \theta_t) \label{eq:mirror_loop}
\end{align*}
setting the gradient w.r.t. $\theta$ to zero, we have
\begin{align*}
\eta  \nabla \mathcal{L}(\theta_{t}) + \nabla \phi(\theta_{t+1}) - \nabla \phi(\theta_t) = 0,
\end{align*}
which when rearranged yields
\begin{align*}
\nabla \phi(\theta_{t+1})  =  \nabla \phi(\theta_t) - \eta  \nabla \mathcal{L}(\theta_{t}) \\
\theta_{t+1}  =\nabla \phi^{-1} (  \nabla \phi(\theta_t) - \eta  \nabla \mathcal{L}(\theta_{t})  ).
\end{align*}
In our case
\begin{align*}
\phi(\theta) = \phi_{M}(\theta) = \frac{1}{2} \theta^T M^2 \theta,
\end{align*}
where $M$ is a diagonal matrix. Therefore,
\begin{align*}
\nabla \phi^{-1} = M^{-2}.
\end{align*}

\section{Proof of Theorem \label{thm:rademacher-bound}}
\begin{proof}
It suffices to bound, with high confidence, the difference between the first term of the meta-objective, and the expected Bregman divergence between initializations and solutions on new tasks sampled from the same task distribution. We will obtain such a bound using Rademacher complexity, and the main result will follow from standard applications of Rademacher complexity-based generalisation bounds \citep{bartlett2002rademacher}, along with the observation that $B_\phi(\theta_\ast||\theta_1) \leq \frac{C^2r^2}{2}$. In particular, we analyse the following class:
\begin{equation}
    \mathcal{F} = \{ (\theta_\ast, \theta_1) \mapsto B_{\phi}(\theta_\ast || \theta_1) \, : \, \phi(\theta) =\max_{j \in \mathbb{N}_N} \, \theta M^2_j \theta, \, M_j = \textup{diag}(\vec m_j), \, \|\vec m_j\|_2 \leq C \}.
\end{equation}
We can bound the Rademacher complexity of this class from above by
\begin{align}
    \hat{R}_n(\mathcal{F}) &= \mathbb{E}_{\sigma} \Bigg [ \sup_{B_{\phi} \in \mathcal{F}} \frac{1}{n} \sum_{i=1}^n \sigma_i B_\phi(\theta_\ast^{(i)} || \theta_1^{(i)}) \Bigg ] \\
    &= \mathbb{E}_{\sigma} \Bigg [ \sup_{M_1, ..., M_N} \frac{1}{2n} \sum_{i=1}^n \sigma_i \max_j \, (\theta_\ast^{(i)} - \theta_1^{(i)})^T M^2_j (\theta_\ast^{(i)} - \theta_1^{(i)}) \Bigg ] \\
    &\leq \mathbb{E}_{\sigma} \Bigg [ \sup_{\vec m_1, ..., \vec m_N} \frac{1}{2n} \sum_{j=1}^N \sum_{i=1}^n \sigma_i \langle \vec m^2_j, (\theta_\ast^{(i)} - \theta_1^{(i)})^2 \rangle \Bigg ] \\
    &= \mathbb{E}_{\sigma} \Bigg [ \sup_{\vec m_1, ..., \vec m_N} \frac{1}{2n} \sum_{j=1}^N \langle \vec m^2_j, \sum_{i=1}^n \sigma_i (\theta_\ast^{(i)} - \theta_1^{(i)})^2 \rangle \Bigg ] \\
    &\leq \mathbb{E}_{\sigma} \Bigg [ \frac{NC^2}{2n} \Bigg \| \sum_{i=1}^n \sigma_i (\theta_\ast^{(i)} - \theta_1^{(i)})^2 \Bigg \|_2 \Bigg ] \\
    &\leq \frac{NC^2r^2}{2\sqrt{n}}
\end{align}
where the second inequality comes from Cauchy-Schwarz, and squaring a vector is understood to be a component-wise operation. The third inequality arises from a well known sequence of steps used when bounding the expected norm of a Rademacher sum---see, e.g., the proof of Lemma 26.10 in \citet{shalev2014understanding}.
\end{proof}

\section{Gradient Computation \label{sec:hyper_gradient}}
We have discussed the parameterisation for the learnable Bregman divergence which eliminates the mirror loop optimisation problem by introducing a closed form solution. As a result, the trilevel optimisation problem is simplified as a bilevel optimisation: 
\begin{align*}
  \min_{\phi}& \mathcal{E}(\theta^\ast(\phi))\\
  \text{s.t.}&\, \theta^\ast(\phi) = \argmin_{\theta} \mathcal{L}_{tr}(\theta) = (\pi_{\phi} \circ \pi_{\phi} ...  \circ \pi_{\phi})(\theta_1)\\
  &\, \pi_{\phi}(\theta_{t}) = \theta_t - \eta M^{-2} \nabla_{\theta} \mathcal{L}_{tr}(\theta_{t}).
\end{align*}
The gradient of the second term in the proposed meta-objective in Eq.\ref{eq:outer_obj} is easy to compute while the first term with respect to $\phi$ is expressed as: 
\begin{align}
\frac{\partial B_{\phi}(\theta_*, \theta_1)}{\partial \phi} & \approx \frac{\partial B_{\phi}(\theta_T, \theta_1)}{\partial \phi} \nonumber\\
&  = \underbrace{\frac{\partial B_{\phi}(\theta_T, \theta_1)}{\partial \phi}}_{\text{direct grad}} 
                          + \underbrace{\frac{\partial B_{\phi}(\theta_T, \theta_1)}{\partial \theta_T} \frac{\partial \theta_T}{\partial \phi}}_{\text{indirect grad}}   \label{eq:direct_indirect_grad}
\end{align}
when T is large enough to satisfy that $\theta_* \approx \theta_T$. The computation of the direct gradient can be easily solved by the existing auto-differentiation library. The indirect grad in Eq~\ref{eq:direct_indirect_grad}, usually termed hypergradient, is much more computationally chanllenging as it is expressed in the form: 
\begin{align}
    \frac{\partial \theta_T}{\partial \phi} & = \sum_{t=1}^T \left(\prod_{t'=t+1}^T A_{t'} \right) B_t \label{ab_closed_form}\\
    \text{s.t.}\, A_{t} & = \frac{\partial \pi_\phi(\theta_{t-1})}{\partial\theta_{t-1}} = I - \eta M^{-2} \frac{\partial^2}{\partial \theta^2}\mathcal{L}_{tr}(\theta_{t-1}), \nonumber \\
 B_t & = \frac{\partial \pi_\phi(\theta_{t-1})}{\partial\phi} = 2 \eta \frac{\partial}{\partial \theta}\mathcal{L}_{tr}(\theta_{t-1})M^{-3}. \nonumber
\end{align}
where we also give the closed-form solution in our setting 
Forward-Mode Differentiation (FMD) and Reverse-Mode Differentiation \cite{franceschi2017forward} are two algorithms to compute Eq~\ref{ab_closed_form}. RMD computes the gradient from the last to the initial step, requiring one to store the entire optimisation trajectory in memory. Thus it is not suitable for our $\phi$ parameterisation whose dimension is $\mathbb{R}^{N\times \kappa}$ where $\kappa$ denotes the number of parameters in the base model which is also the number of elements on the diagonal on $M_i$. In comparison, FMD updates the hypergradient in parallel with in inner loop optimisation by:
\begin{align*}
\frac{\partial \theta_t}{\partial \phi }&= \frac{\partial \pi(\theta_{t-1})}{\partial \theta_{t-1}} \frac{\partial \theta_{t-1}}{\partial \phi} + \frac{\partial \pi(\theta_{t-1})}{\partial \phi }, \\
\end{align*}
where it only requires the information from step $t-1$.
\section{Training on linear model\label{sec:linear}}
\begin{figure}[t]
\label{fig:linear_romnist}
\centering
    \includegraphics[width=0.48\linewidth]{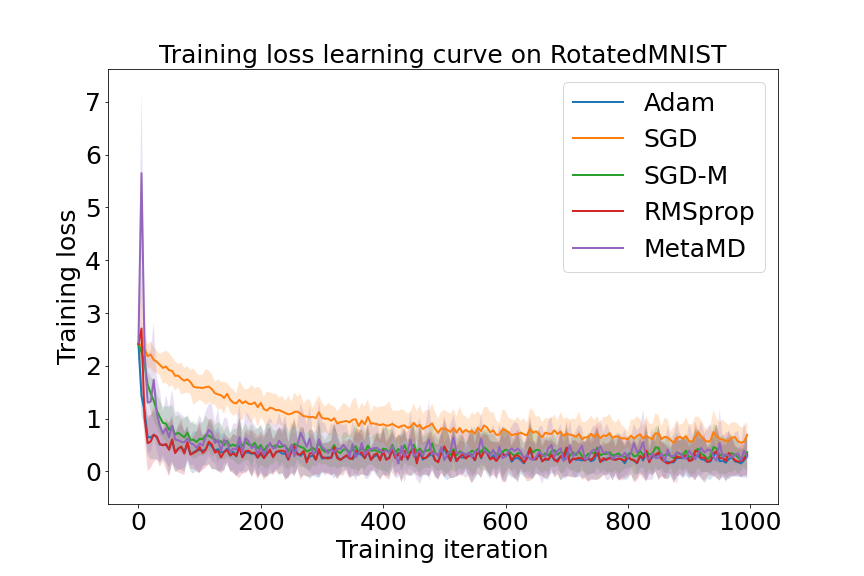}
\caption{Average Learning curve comparison of the training loss over different domains on RotatedMNIST produced by different optimisers. The loss learning curve of the different optimisers training on linear model with 1000 training iterations.}

\label{fig:rmnist_learning_curve}
\end{figure}

We study the convex setting when the base model is linear. In Table~\ref{tab:rotatedmnist_linear}, we can see that in this setting all the optimisers have very similar performance due to the single global minima caused by convexity, but converge at different speeds shown in Fig \ref{fig:linear_romnist}. SGD converges in a slower ratio than others. 
\begin{table*}[t]
\centering
\label{tab:rotatedmnist_linear}
\caption{Test Accuracy (\%) on RotatedMNIST with linear model} 
\resizebox{0.9\textwidth}{!}{
\begin{tabular}{ l | l | c c c c c c}
\toprule
 &Method & 0 & 15 & 30 & 45 & 60 & 75 \\
\cmidrule{1-8}
\multirow{6}{*}{\rotatebox[origin=c]{90}{\textbf{Linear Model}}}
&SGD           & 86.12 $\pm$ 0.80 & 86.89 $\pm$ 0.34 & 86.38 $\pm$ 0.39 & 86.14 $\pm$ 0.45 & 86.85 $\pm$ 0.54 & 85.39 $\pm$ 0.31 \\
&SGD-M         & 87.57 $\pm$ 0.44 & 87.62 $\pm$ 0.18 & 87.58 $\pm$ 0.33 & 87.13 $\pm$ 0.55 & 87.52 $\pm$ 0.48 & 87.95 $\pm$ 0.23 \\
&Adam          & 87.51 $\pm$ 0.49 & 87.29 $\pm$ 0.56 & 87.83 $\pm$ 0.57 & 87.84 $\pm$ 0.94 & 87.68 $\pm$ 0.34 & 87.93 $\pm$ 0.69 \\
&RMSprop       & 87.34 $\pm$ 0.77 & 87.47 $\pm$ 0.61 & 87.16 $\pm$ 1.04 & 87.24 $\pm$ 1.00 & 87.46 $\pm$ 0.77 & 87.64 $\pm$ 0.53 \\
\cmidrule{2-8}
&\ourAcronym{} & 87.57 $\pm$ 0.68 & 87.58 $\pm$ 0.52 & 87.77 $\pm$ 0.64 & 87.54 $\pm$ 0.40 & 87.68 $\pm$ 0.35 & 87.93 $\pm$ 0.29   \\
\cmidrule{1-8}
\cmidrule{1-8}
\end{tabular}
}
\end{table*}
\section{Hyperparameter Tuning\label{sec:hyper_tuning_shmodel}}
\keypoint{Grid Search} For tuning the hyperparameters in linear in Appendix~\ref{sec:linear} and 3-layer MLPs~\ref{sec:rotatedMNIST_MLP} model settings, we sweep over the learning rates $\{0.1, 0.05, 0.01, 0.005, 0.001\}$ and weight decay parameters of $\{ 0.001, 0.0001, 0.0005\}$ for the SGD, SGD-M and RMSprop. In terms of Adam, we do grid search over the learn rates $\{0.3, 0.2, 0.1, 0.01, 0.001\}$ and weight decay $\{0.001, 0.0001, 0.0005\}$. 

\keypoint{Bayesian Optimisation} We implement our BayesOpt using \citet{balandat2020botorch}. The model the expected performance using a Gaussian process with RBF kernel, which maps the learning rate and weight decay to the estimated validation accuracy. This also provides uncertainty information to the Upper Confidence Bound (UCB) acquisition function for exploring/exploiting the hyperparameter space. For each model selection in the meta-test stage, we run the Bayesian optimisation for 25 iterations.

\section{Training loss learning curve for DiverseDigits dataset \label{sec:whole_lenet_mnist}}
We give all the training loss learning curves on DiverseDigits in Fig~\ref{fig:whole_mnist_lenet}. It can be noticed that the conclusion we drew that \ourAcronym{} is clearly faster than SGD and SGD-M in training convergence in Section~\ref{sec:rotatedMNIST_MLP} is further supported.
\cut{
\begin{figure}[tb]
    \centering
    \includegraphics[width=0.3\linewidth]{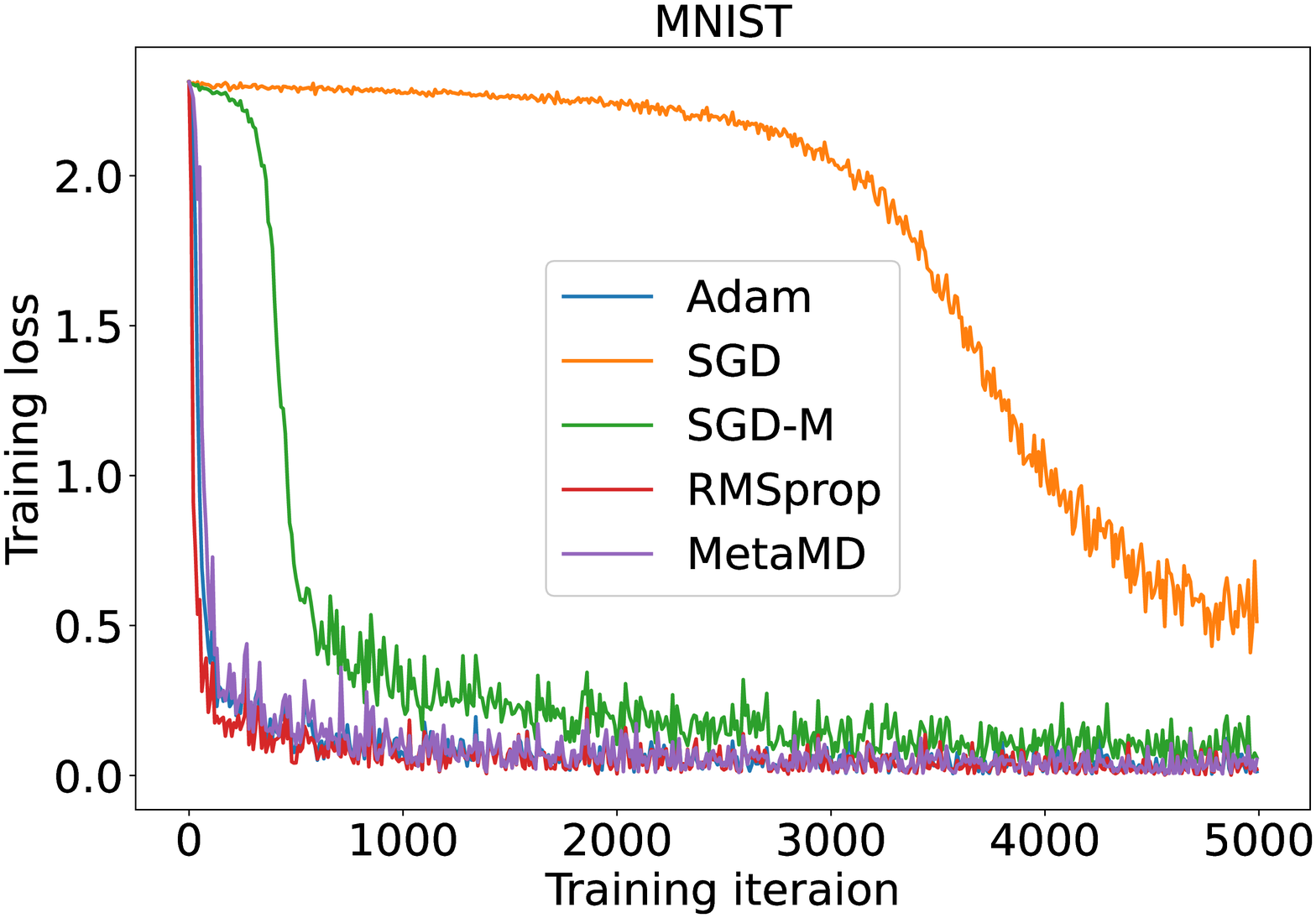}
    \includegraphics[width=0.3\linewidth]{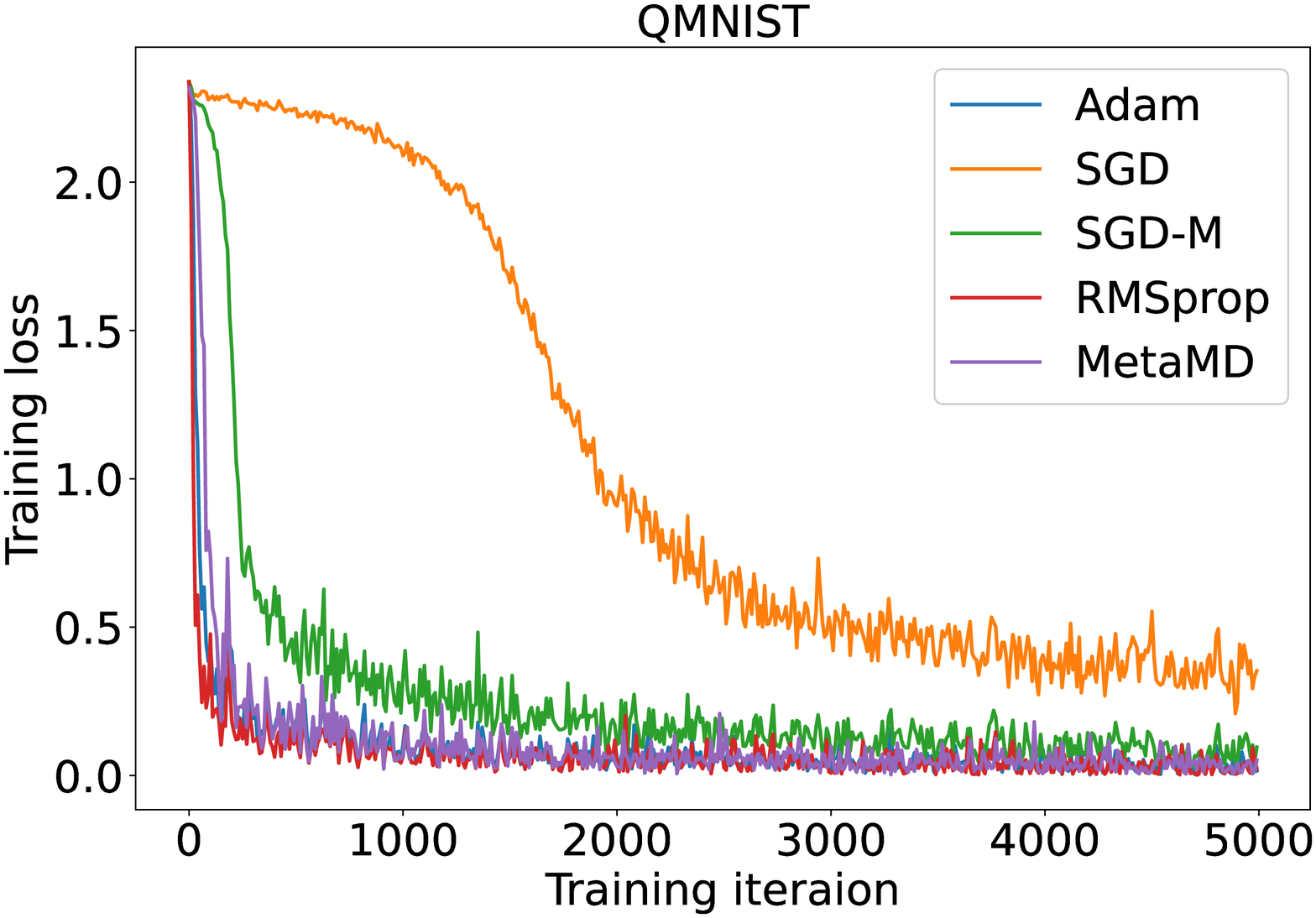}
    \includegraphics[width=0.3\linewidth]{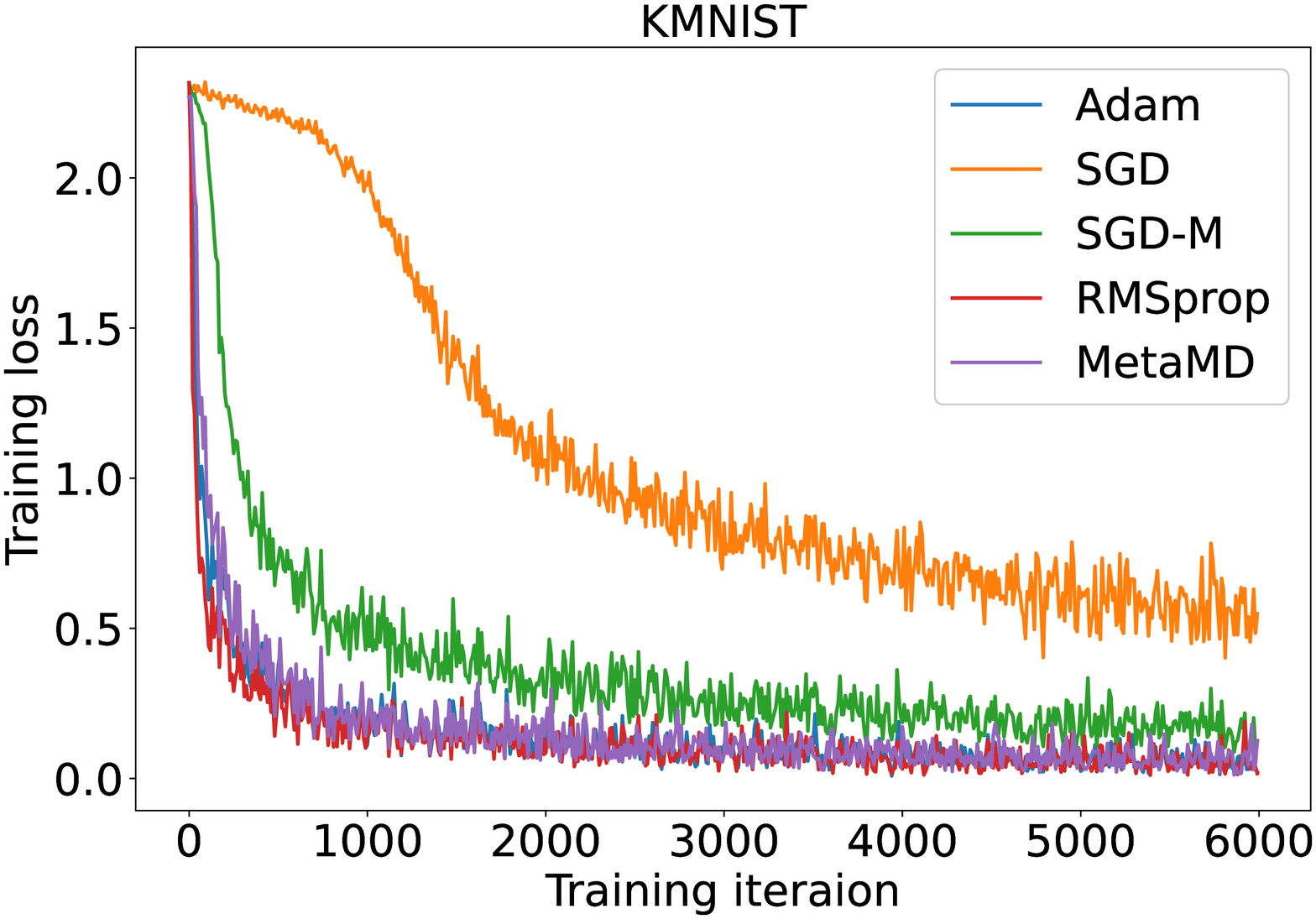}
    \includegraphics[width=0.3\linewidth]{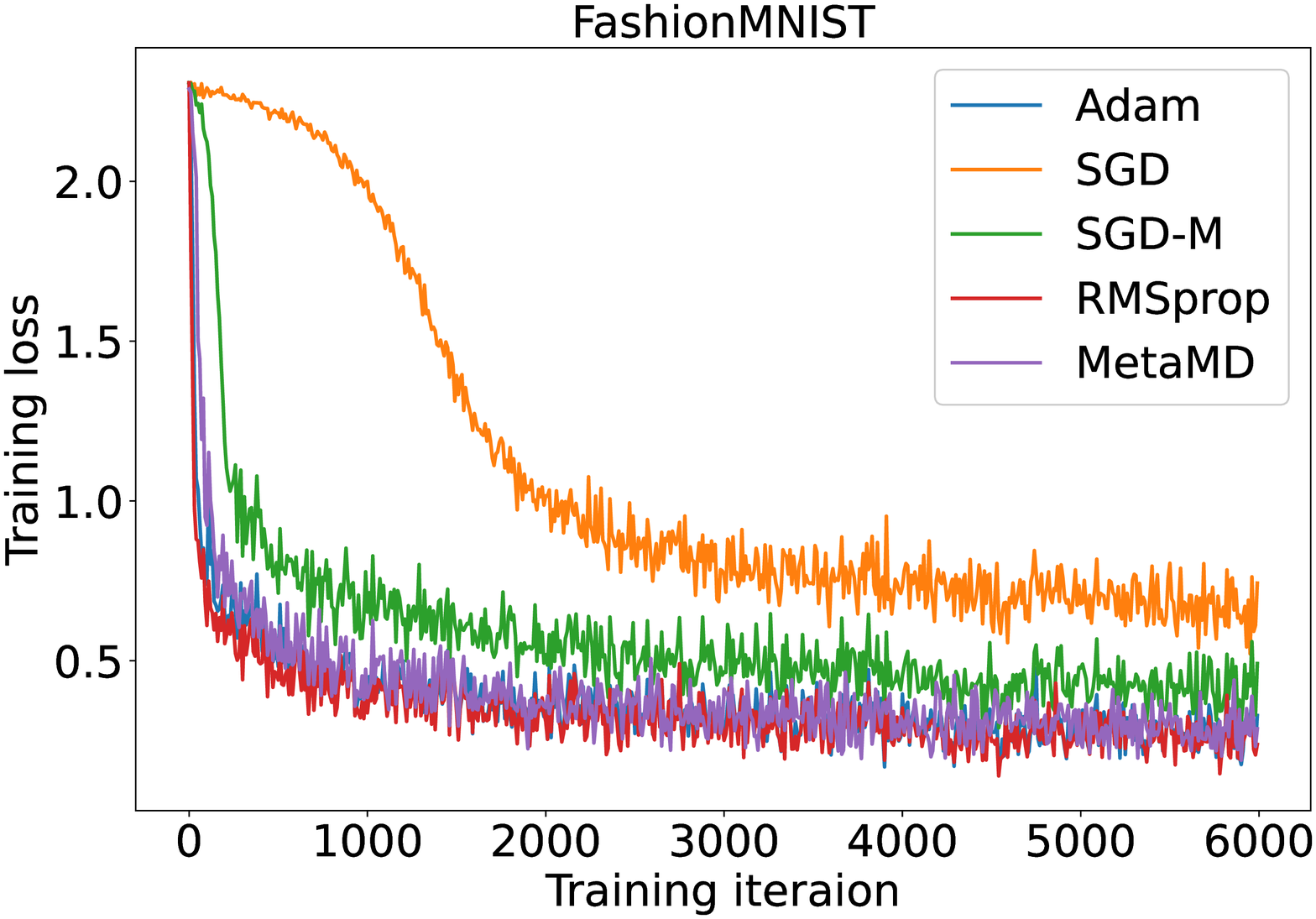}
    \includegraphics[width=0.3\linewidth]{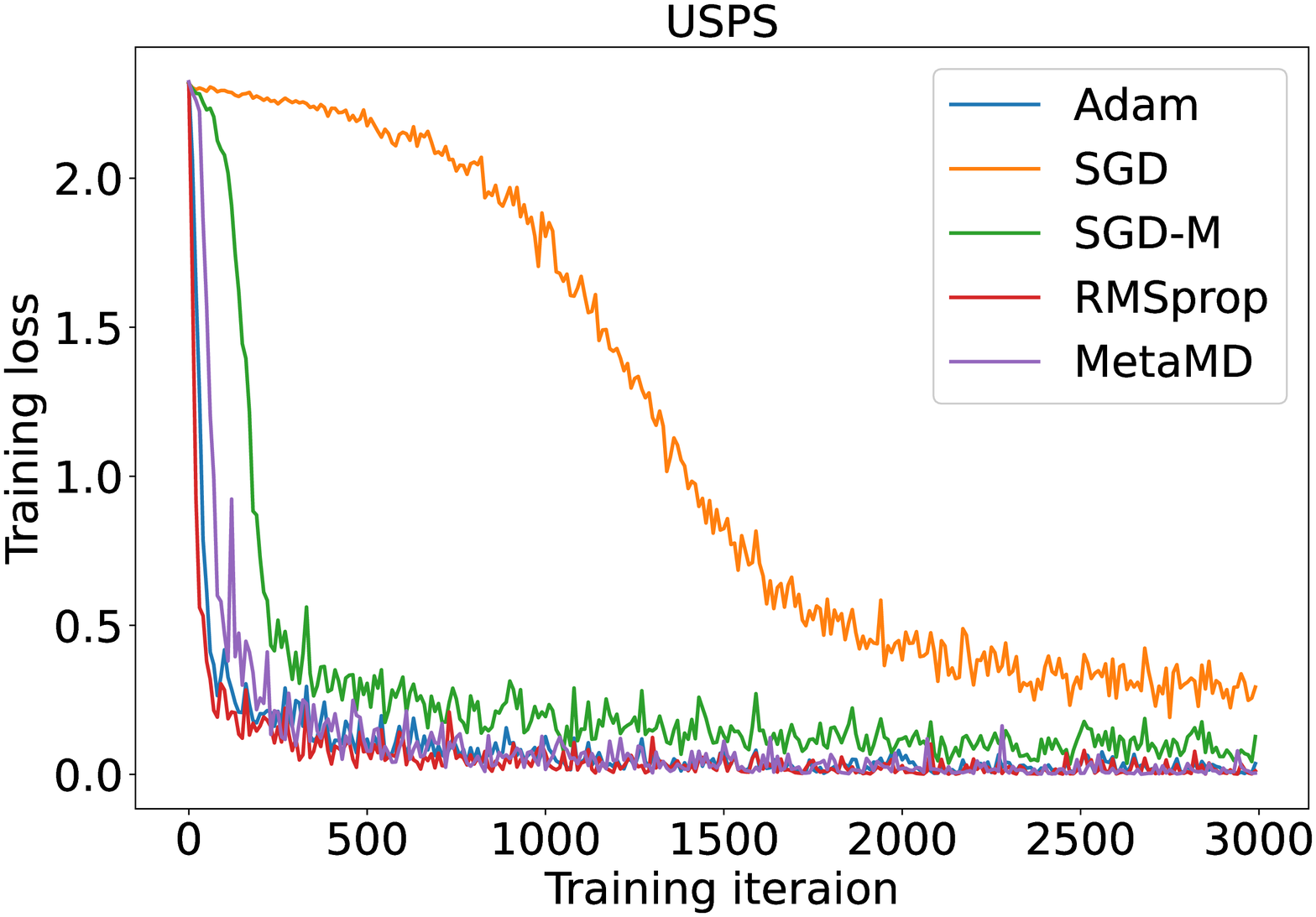}
    \includegraphics[width=0.3\linewidth]{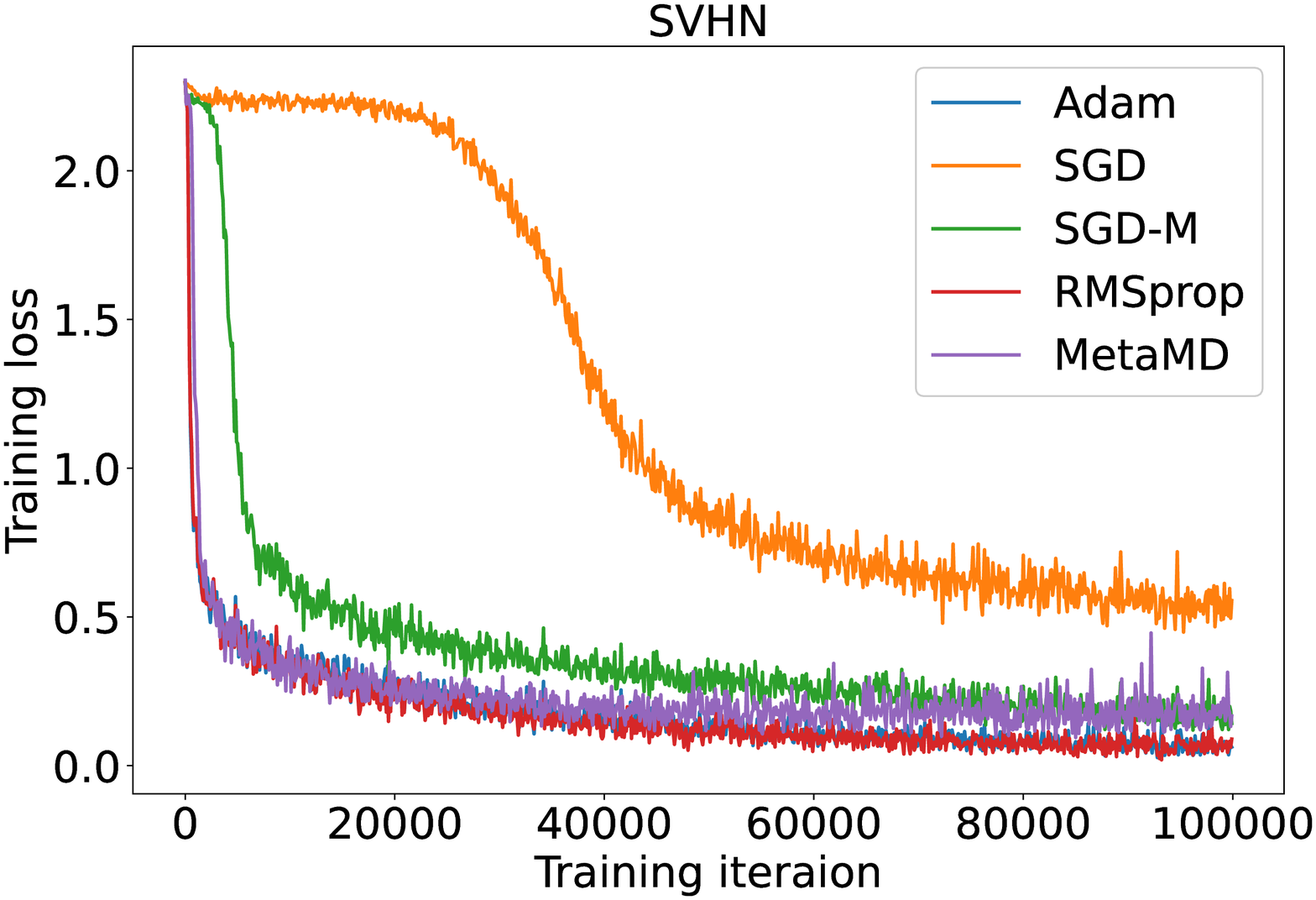}
\caption{Convergence comparison of different optimisers on DiverseDigits.}
\label{fig:whole_mnist_lenet}
\end{figure}
}
\begin{figure}[tb]
    \centering
    \includegraphics[width=0.3\linewidth]{mnistf_learning_curve/learning_curve_MNIST.pdf}
    \includegraphics[width=0.3\linewidth]{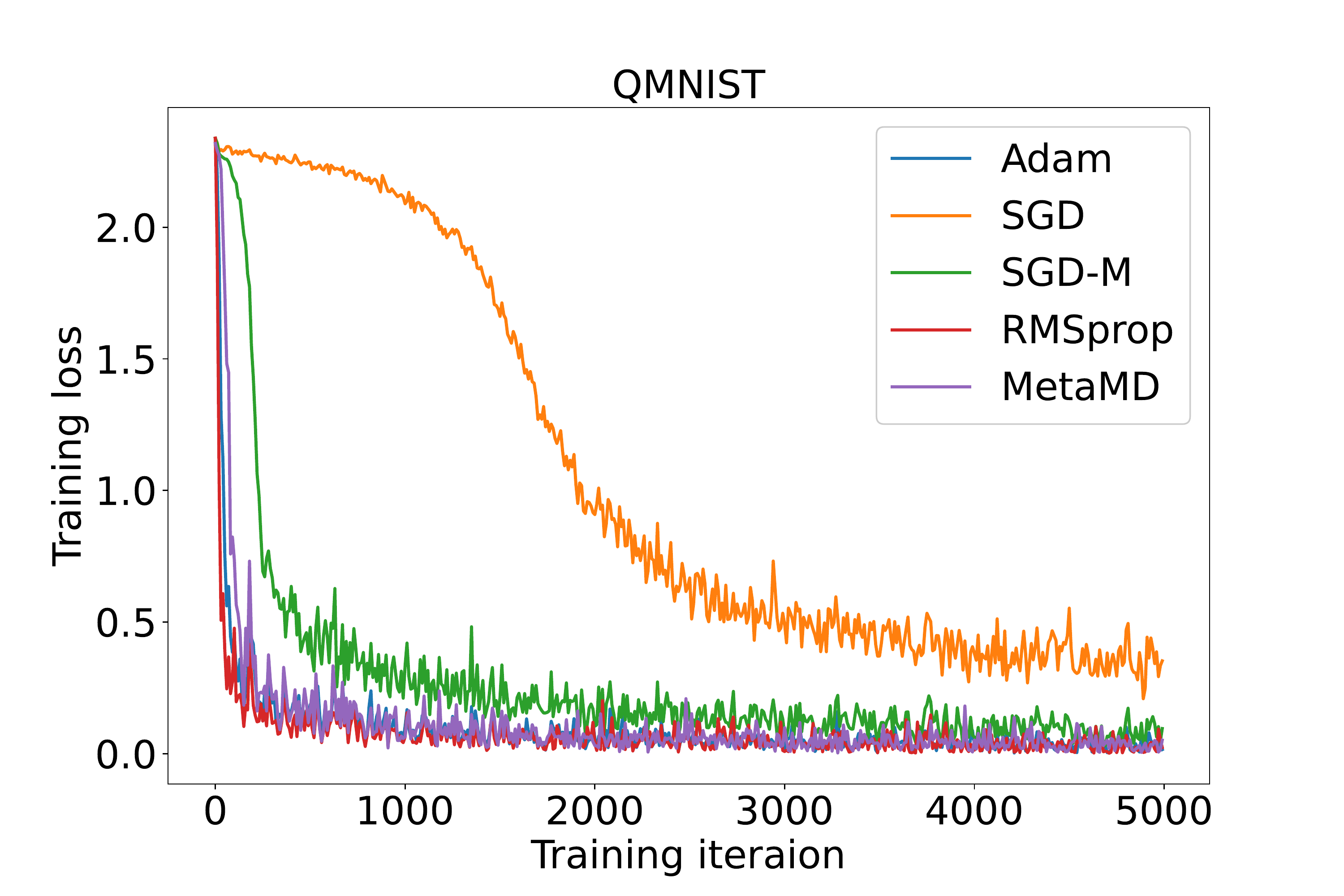}
    \includegraphics[width=0.3\linewidth]{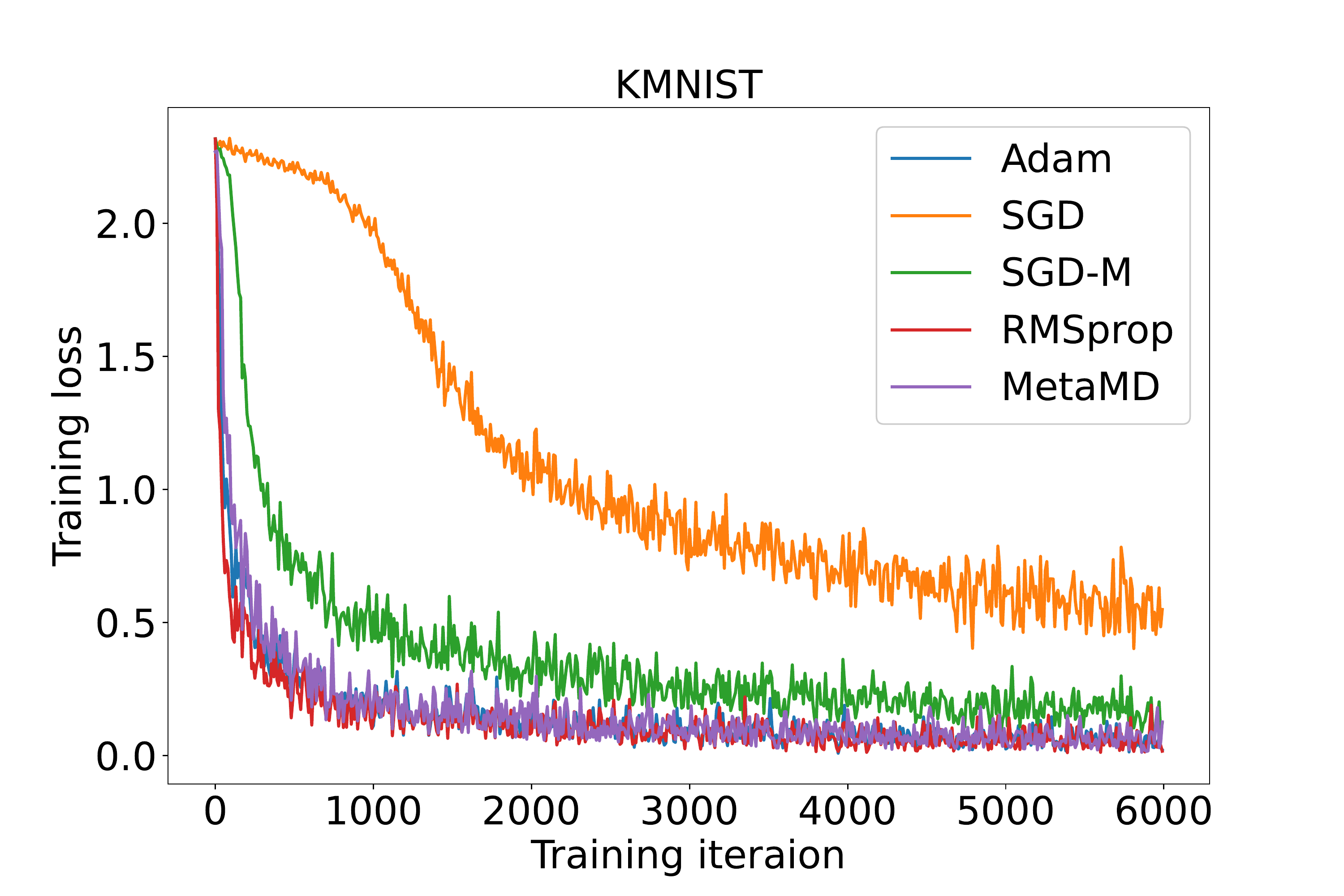}
    \includegraphics[width=0.3\linewidth]{mnistf_learning_curve/learning_curve_FashionMNIST.pdf}
    \includegraphics[width=0.3\linewidth]{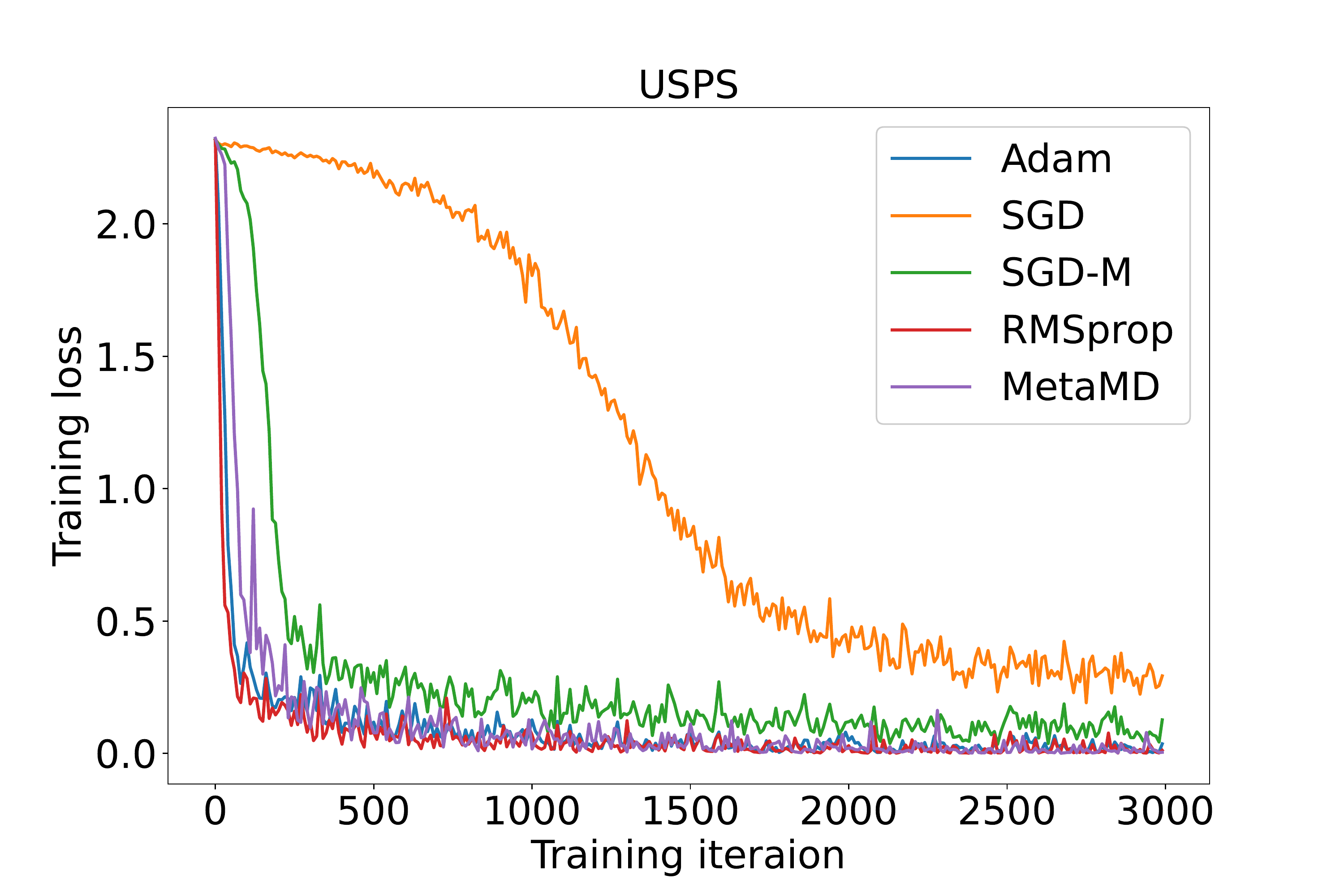}
    \includegraphics[width=0.3\linewidth]{mnistf_learning_curve/learning_curve_SVHN.pdf}
\caption{Convergence comparison of different optimisers on DiverseDigits.}
\label{fig:whole_mnist_lenet}
\end{figure}

\end{document}